\documentclass{article} % For LaTeX2e
\usepackage{iclr2024_conference,times}

\usepackage{amsmath,amsfonts,bm}

\def\figref#1{figure~\ref{#1}}
\def\Figref#1{Figure~\ref{#1}}

\def\secref#1{section~\ref{#1}}
\def\Secref#1{Section~\ref{#1}}
\def\eqref#1{equation~\ref{#1}}
\def\algref#1{algorithm~\ref{#1}}
\def\1{\bm{1}}

\DeclareMathAlphabet{\mathsfit}{\encodingdefault}{\sfdefault}{m}{sl}
\SetMathAlphabet{\mathsfit}{bold}{\encodingdefault}{\sfdefault}{bx}{n}

\DeclareMathOperator*{\argmin}{arg\,min}

\usepackage{etoolbox}

\usepackage{hyperref}
\usepackage{url}

\usepackage{booktabs}       % professional-quality tables

\usepackage{multirow}
\usepackage{graphicx}
\usepackage{wrapfig}
\usepackage[labelformat=simple]{subcaption}

\usepackage{caption}
\usepackage{bbm}

\usepackage{selectp}
\usepackage{siunitx}
\DeclareSIUnit{\percent}{\%}
\DeclareSIUnit{\gigabyte}{GB}

\usepackage{amsmath}

\usepackage{comment}

\usepackage{amssymb}

\let\emptyset\varnothing

\usepackage{algorithm}
\usepackage{algorithmic}

\usepackage{abbr_style} % own helper for common abbreviations
\renewcommand{\secref}[1]{Section~\ref{#1}}
\renewcommand{\eqref}[1]{Equation~\ref{#1}}
\renewcommand{\figref}[1]{Figure~\ref{#1}}
\newcommand{\tabref}[1]{Table~\ref{#1}}

\renewcommand{\algref}[1]{Algorithm~\ref{#1}}

\newcommand{\appendixref}[1]{Appendix~\ref{#1}} % helpers for referencing

\definecolor{cadmiumgreen}{rgb}{0.0, 0.42, 0.24}
\newcommand{\oursc}{LDCE-cls}
\newcommand{\ourst}{LDCE-txt}

\title{Latent Diffusion \\Counterfactual Explanations}

\author{Karim Farid\thanks{Equal contribution.}\hspace{1cm}Simon Schrodi$^{*}$\hspace{1cm}Max Argus\hspace{1cm}Thomas Brox\\
University of Freiburg\\
\texttt{\{faridk,schrodi,argusm,brox\}@cs.uni-freiburg.de}
}

\iclrpreprintcopy

\newtoggle{iclrsubmission}
\toggletrue{iclrsubmission}

\ificlrfinal
\togglefalse{iclrsubmission}
\fi

\ificlrpreprint
\togglefalse{iclrsubmission}
\fi

\newcommand{\codeURL}{\url{https://anonymous.4open.science/r/ldce}}
\ifboolexpr{not togl {iclrsubmission}}{
\renewcommand{\codeURL}{\url{https://github.com/lmb-freiburg/ldce}}
}

\begin{document}

\maketitle

% Keywords: Counterfactual Explanations, Diffusion Models, Explainable AI
%TL;DR: We present a method for generating visual counterfactual explanations using a foundational latent diffusion model, augmented by a novel consensus guidance mechanism.
\begin{abstract}
Counterfactual explanations have emerged as a promising method for elucidating the behavior of opaque black-box models. Recently, several works leveraged pixel-space diffusion models for counterfactual generation. To handle noisy, adversarial gradients during counterfactual generation--causing unrealistic artifacts or mere adversarial perturbations--they required either auxiliary adversarially robust models or computationally intensive guidance schemes. However, such requirements limit their applicability, e.g., in scenarios with restricted access to the model's training data. To address these limitations, we introduce Latent Diffusion Counterfactual Explanations (LDCE). LDCE harnesses the capabilities of recent class- or text-conditional foundation latent diffusion models to expedite counterfactual generation and focus on the important, semantic parts of the data. Furthermore, we propose a novel consensus guidance mechanism to filter out noisy, adversarial gradients that are misaligned with the diffusion model's implicit classifier. We demonstrate the versatility of LDCE across a wide spectrum of models trained on diverse datasets with different learning paradigms. Finally, we showcase how LDCE can provide insights into model errors, enhancing our understanding of black-box model behavior.
\end{abstract}

\section{Introduction}
Deep learning systems achieve remarkable results across diverse domains (\eg, \citet{brown2020language,jumper2021highly}), yet their opacity presents a pressing challenge: as their usage soars in various applications, it becomes increasingly important to understand their underlying inner workings, behavior, and decision-making processes \citep{arrieta2020explainable}.
There are various lines of work that facilitate a better understanding of model behavior, including: pixel attributions (\eg, \citet{simonyan2013deep,bach2015pixel,selvaraju2017grad,lundberg2017unified}), feature visualizations (\eg, \citet{erhan2009visualizing,simonyan2013deep,olah2017feature}), concept-based methods (\eg, \citet{bau2017network,kim2018interpretability,koh2020concept}), inherently interpretable models (\eg, \citet{brendel2018approximating,chen2019looks,bohle2022b}), and counterfactual explanations (\eg, \citet{wachter2017counterfactual,goyal2019counterfactual}).

In this work, we focus on counterfactual explanations that modify a (f)actual input with the \emph{smallest semantically meaningful} change such that a target model changes its output. 
Formally, given a (f)actual input \begin{math}x^F\end{math} and (target) model \begin{math}f\end{math}, \citet{wachter2017counterfactual} proposed to find the counterfactual explanation \begin{math}x^{\textrm{CF}}\end{math} that achieves a desired output \begin{math}y^{\textrm{CF}}\end{math} defined by loss function \begin{math}\mathcal{L}\end{math} and stays as close as possible to the (f)actual input defined by a distance metric \begin{math}d\end{math}, as follows:
\begin{equation}\label{eq:cf}
    x^{\textrm{CF}}\in\argmin\limits_{x'} \lambda_c \mathcal{L}(f(x'),\;y^{\textrm{CF}})+ \lambda_d d(x',\;x^{\textrm{F}}) \quad .
\end{equation}
Generating (visual) counterfactual explanations from above optimization problem poses a challenge since, \eg, relying solely on the classifier gradient often results in adversarial examples rather than counterfactual explanations with semantically meaningful (\ie, human comprehensible) changes.
Thus, previous works resorted to adversarially robust models (\eg, \citet{santurkar2019image,boreiko2022sparse}), restricted the set of image manipulations (\eg, \citet{goyal2019counterfactual,wang2021imagine}), used generative models (\eg, \citet{samangouei2018explaingan,lang2021explaining,khorram2022cycle,jeanneret2022diffusion}), or used mixtures of aforementioned approaches (\eg, \citet{augustin2022diffusion}) to regularize towards the (semantic) data manifold.
However, these requirements or restrictions can hinder the applicability, \eg, in real-world scenarios with restricted data access due to data privacy reasons.

Thus, we introduce Latent Diffusion Counterfactual Explanations~(LDCE) which is devoid from such limitations. 
LDCE leverages recent class- or text-conditional foundational diffusion models combined with a novel \emph{consensus guidance mechanism} that filters out adversarial gradients of the target model that are not aligned with the gradients of the diffusion model's implicit classifier. Moreover, through the decoupling of the semantic from pixel-level details by \emph{latent} diffusion models \citep{rombach2022high}, we not only expedite counterfactual generation but also disentangle semantic from pixel-level changes during counterfactual generation. To the best of our knowledge, LDCE is the first counterfactual approach that can be applied to \emph{any} classifier; independent of the learning paradigm (\eg, supervised or self-supervised) and with extensive domain coverage (as wide as the foundational model's data coverage), while generating high-quality visual counterfactual explanations of the tested classifier; see \Figref{fig:teaser}. Code is available at \codeURL.

In summary, our key contributions are the following:
\begin{itemize}
    \item By leveraging recent class- or text-conditional foundation diffusion models \citep{rombach2022high}, we present the first approach that is both \emph{model-} and \emph{dataset-agnostic} (restricted only by the domain coverage of the foundation model).
    \item We introduce a novel \emph{consensus guidance mechanism} that eliminates confounding elements, such as an auxiliary classifier, from the counterfactual generation by leveraging foundation models' implicit classifiers as a filter to ensure semantically meaningful changes in the counterfactual explanations of the tested classifiers.
\end{itemize}

\newcommand{\toar}{$\rightarrow$~}
\begin{figure}
    \centering
    \begin{subfigure}[b]{0.35\linewidth}
        \includegraphics[width=\linewidth]{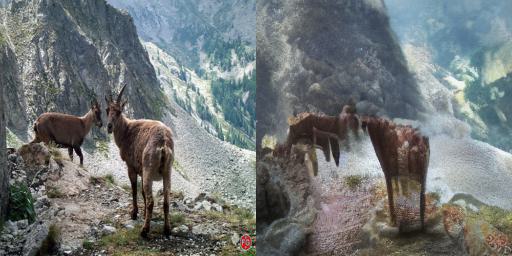}
        \caption{\centering alp \toar coral reef \newline on ImageNet with ResNet-50}
        \label{fig:teaser_imagenet}
    \end{subfigure}
    \hspace{1cm}
    \begin{subfigure}[b]{0.35\linewidth}
        \includegraphics[width=\linewidth]{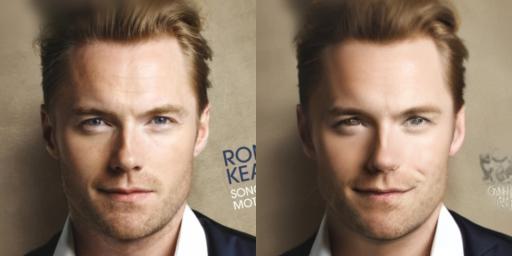}
        \caption{\centering no-smile \toar smile \newline  on CelebA HQ with DenseNet-121}
        \label{fig:teaser_celeb}
    \end{subfigure}
    \begin{subfigure}[b]{0.35\linewidth}
        \includegraphics[width=\linewidth]{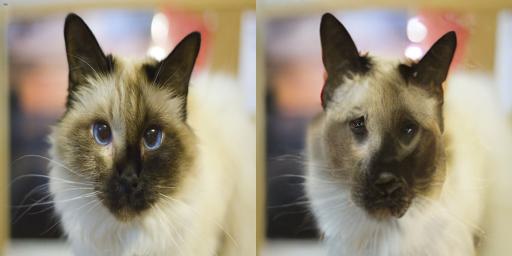}
        \caption{\centering birman \toar American pit bull \newline  on Oxford Pets with OpenCLIP}
    \end{subfigure}
    \hspace{1cm}
    \begin{subfigure}[b]{0.35\linewidth}
        \includegraphics[width=\linewidth]{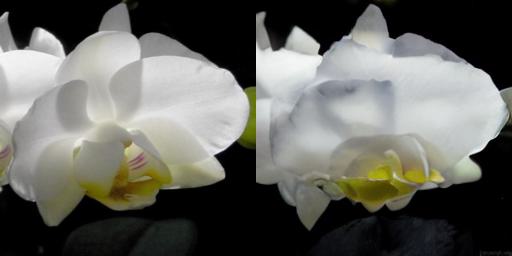}
        \caption{\centering moon orchid \toar rose \newline on Oxford Flowers with DINO+linear}
        \label{fig:teaser_flower}
    \end{subfigure}
    \caption{LDCE(-txt) can be applied to \emph{any} classifier, is \emph{dataset-agnostic}, and works across various learning paradigms. We show counterfactual explanations (right) for the original image (left) for various dataset and classifier combinations.
    }
    \label{fig:teaser}
\end{figure}

\section{Background}
\subsection{Diffusion models}
Recent work showed that diffusion models can generate high-quality images \citep{sohl2015deep,song2019generative,ho2020denoising,song2021denoising,rombach2022high}. The main idea is to gradually add small amounts of Gaussian noise to the data in the forward diffusion process and gradually undoing it in the learned reverse diffusion process. Specifically, given scalar noise scales \begin{math}\{\alpha_t\}_{t=1}^{T}\end{math} and an initial, clean image \begin{math}x_0\end{math}, the forward diffusion process generates intermediate noisy representations \begin{math}\{x_t\}_{t=1}^T\end{math}, with \begin{math}T\end{math} denoting the number of time steps. We can compute \begin{math}x_t\end{math} by
\begin{equation}
    \label{eq:forward_diffusion}
    x_t=\sqrt{\alpha_t}x_0 + \sqrt{1-\alpha_t}\epsilon_t,\quad\textrm{where}\quad\epsilon_t\sim\mathcal{N}(\mathbf{0},\mathbf{I})\qquad .
\end{equation}
The score estimator (\ie, parameterized denoising network) \begin{math}\epsilon_{\theta}(x_t,\;t)\end{math}--typically a modified U-Net \citep{ronneberger2015unet}--learns to undo the forward diffusion process for a pair \begin{math}(x_t,\;t)\end{math}:
\begin{equation}\label{eq:denoising}
    \epsilon_{\theta}(x_{t},\;t)\approx \hat{\epsilon}_t = \frac{x_t-\sqrt{\alpha_t}x_0}{\sqrt{1-\alpha_t}} \quad .
\end{equation}
Note that by rewriting \eqref{eq:denoising} (or \ref{eq:forward_diffusion}), we can approximately predict the clean data point
\begin{equation}
    \label{eq:DDIM_x0}
    \hat{x}_{0}\approx\frac{x_t-\sqrt{1-\alpha_t}\epsilon_{\theta}(x_t,\;t)}{\sqrt{\alpha_t}}\quad .
\end{equation}
To gradually denoise, we can sample the next less noisy representation \begin{math}x_{t-1}\end{math} with a sampling method \begin{math}S(x_t,\;\hat{\epsilon}_t,\;t)\rightarrow x_{t-1}\end{math}, such as the DDIM sampler \citep{song2021denoising}:
\begin{equation}\label{eq:ddim_sampler}
    x_{t-1}=\sqrt{\alpha_{t-1}}\frac{x_t-\sqrt{1-\alpha_t}\hat{\epsilon}_{t}}{\sqrt{\alpha_t}}+\sqrt{1-\alpha_{t-1}-\sigma^2_t}\hat{\epsilon}_{t}+\sigma_t\epsilon_t\quad .
\end{equation}

\paragraph{Latent diffusion models}~
In contrast to GANs \citep{goodfellow2020generative}, VAEs \citep{kingma2014auto,rezende2014stochastic}, or normalizing flows \citep{rezende2015variational}, (pixel-space) diffusion models' intermediate representations are high-dimensional, rendering the generative process computationally very intensive. To mitigate this, \citet{rombach2022high} proposed to operate diffusion models in a perceptually equivalent, lower-dimensional latent space \begin{math}\mathcal{Z}\end{math} of a regularized autoencoder \begin{math}\mathcal{A}(x)=\mathcal{D}(\mathcal{E}(x))\approx x\end{math} with encoder \begin{math}\mathcal{E}\end{math} and decoder \begin{math}\mathcal{D}\end{math} \citep{esser2021taming}. Note that this also decouples semantic from perceptual compression s.t. the ``focus [of the diffusion model is] on the important, semantic bits of the data" (\citet{rombach2022high}, p. 4).

\paragraph{Controlled image generation}~
To condition the generation by some condition \begin{math}c\end{math}, a class label or text, we need to learn a score function \begin{math}\nabla_x \log p(x|c)\end{math}. Through, Bayes' rule we can decompose the score function into an unconditional and conditional component: 
\begin{equation}\label{eq:guidance}
    \nabla_x \log p_\eta(x|c)=\nabla_x \log p(x)+\eta\nabla_x \log p(c|x) \quad ,
\end{equation}
where the guidance scale \begin{math}\eta\end{math} governs the influence of the conditioning signal.
Note that \begin{math}\nabla_x \log p(x)\end{math} is just the unconditional score function and \begin{math}\nabla_x \log p(c|x)\end{math} can be a standard classifier. However, intermediate representations of the diffusion process have high noise levels and directly using a classifier's gradient may result in mere adversarial perturbations \citep{augustin2022diffusion}. To overcome this, previous work used noise-aware classifiers \citep{dhariwal2021diffusion}, optimized intermediate representation of the diffusion process \citep{jeanneret2023adversarial,wallace2023end}, or used one-step approximations \citep{avrahami2022blended,augustin2022diffusion,bansal2023universal}. 
In contrast to these works, \citet{ho2022classifier} trained a conditional diffusion model \begin{math}\nabla_x \log p(x|c)\end{math} with conditioning dropout and leveraged Bayes' rule, \ie,
\begin{equation}\label{eq:implicit}
    \nabla_x \log p(c|x)=\nabla_x \log p(x|c)-\nabla_x \log p(x) \quad ,
\end{equation}
to substitute the conditioning component \begin{math}\nabla_x \log p(c|x)\end{math} from \eqref{eq:guidance}:
\begin{equation}
    \nabla_x \log p_\eta(x|c)=\nabla_x \log p(x)+\eta(\nabla_x \log p(x|c)-\nabla_x \log p(x)) \quad .
\end{equation}

\subsection{Counterfactual explanations}\label{sub:rw_ce}
A \underline{c}ounter\underline{f}actual explanation \begin{math}x^{\textrm{CF}}\end{math} is a sample with the \emph{smallest} and \emph{semantically meaningful} change to an original \underline{f}actual input \begin{math}x^{\textrm{F}}\end{math} in order to achieve a \emph{desired output}. In contrast to adversarial attacks, counterfactual explanations focus on semantic (\ie, human comprehensible) changes.
Initial works on visual counterfactual explanations used gradient-based approaches \citep{wachter2017counterfactual,santurkar2019image,boreiko2022sparse} or restricted the set of image manipulations \citep{goyal2019counterfactual,akula2020cocox,wang2021imagine,van2021interpretable,vandenhende2022making}.
Other works leveraged invertible networks \citep{hvilshoj2021ecinn}, deep image priors \citep{thiagarajan2021designing}, or used generative models to regularize towards the image manifold to generate high-quality visual counterfactual explanations \citep{samangouei2018explaingan,lang2021explaining,sauer2021counterfactual,rodriguez2021beyond,khorram2022cycle,jacob2022steex}. Recent works also adopted (pixel-space) diffusion models due to their remarkable generative capabilities \citep{sanchez2022diffusion,jeanneret2022diffusion,jeanneret2023adversarial,augustin2022diffusion}.
\section{Latent diffusion counterfactual explanations (LDCE)}\label{sec:ldvce}
\begin{algorithm}[t]
\begin{algorithmic}
\STATE \textbf{Input:} (f)actual image \begin{math}x^{\textrm{F}}\end{math}, condition \begin{math}c\end{math}, target model \begin{math}f\end{math}, encoder \begin{math}\mathcal{E}\end{math}, decoder \begin{math}\mathcal{D}\end{math}, sampler \begin{math}S\end{math}, distance function \begin{math}d\end{math}, ``where" function \begin{math}\phi\end{math}, consensus threshold \begin{math}\gamma\end{math}, time steps \begin{math}T\end{math}, weighing factors \begin{math}\eta,\;\lambda_c,~\lambda_d\end{math}
\STATE \textbf{Output:} counterfactual image \begin{math}x^{\textrm{CF}}\end{math}
\STATE \begin{math}z_T\leftarrow \sqrt{\alpha_t}\mathcal{E}(x^{F}) + \sqrt{1-\alpha_t}\epsilon_t,\quad\textrm{where}\quad\epsilon_t\sim\mathcal{N}(\mathbf{0},\mathbf{I})\quad\textrm{// \cf, \eqref{eq:forward_diffusion}}\end{math}
\FOR{\begin{math}t=T,\; \dots,\; 0\end{math}}
	\STATE \begin{math}\epsilon_{uc},\;\epsilon_{c}\leftarrow \epsilon_{\theta}(z_t,\;t,\;\emptyset),\epsilon_{\theta}(z_t,\;t,\;c)\;\end{math} 
	\STATE \begin{math}\hat{x}_0\leftarrow \mathcal{D}(\frac{z_t-\sqrt{1-\alpha_t}\epsilon_{uc}}{\sqrt{\alpha_t}})\quad \textrm{// \cf, \eqref{eq:DDIM_x0}}\end{math}
	\STATE \begin{math}\texttt{cls\_score},\;\texttt{dist\_score}\leftarrow \sqrt{1-\alpha_t} \nabla_{z_t} \mathcal{L}(f(\hat{x}_0),\;c),\;\sqrt{1-\alpha_t} \nabla_{z_t} d(\hat{x}_0, x^{\textrm{F}}) \end{math}
	\STATE \begin{math}\alpha_i\leftarrow \angle( \texttt{cls\_score}_i,  (\epsilon_{c}-\epsilon_{uc})_i)\end{math}
	\STATE \begin{math}\texttt{consensus}\leftarrow\phi(\alpha_i,\;\gamma,\; \texttt{cls\_score}_i,\; \mathbf{0})\end{math}
	\STATE \begin{math}\hat{\epsilon}_t\leftarrow\epsilon_{uc}+\eta\cdot(\lambda_c \frac{\texttt{consensus}}{||\texttt{consensus}||_2} + \lambda_d \frac{\texttt{dist\_score}}{||\texttt{dist\_score}||_2})\cdot ||\epsilon_c||_2\end{math}
	\STATE \begin{math}z_{t-1}\leftarrow S(z_t,\;\hat{\epsilon}_t,\;t)\quad\textrm{// \eg, \eqref{eq:ddim_sampler}}\end{math}
	\ENDFOR
	\STATE \begin{math}x^{\textrm{CF}}\leftarrow \mathcal{D}(z_0)\end{math}
\end{algorithmic}
\caption{Latent diffusion counterfactual explanations (LDCE).}\label{alg:method}
\end{algorithm}
Since generating (visual) counterfactual explanations \begin{math}x^{\textrm{CF}}\end{math} is inherently challenging due to high chance of adversarial perturbation by solely relying on the target model's gradient, recent work resorted to (pixel-space) diffusion models to regularize counterfactual generation towards the data manifold by employing the following two-step procedure:
\begin{enumerate}
    \item \textbf{Abduction}: add noise to the (f)actual image \begin{math}x^{\textrm{F}}\end{math} through the forward diffusion process, and
    \item \textbf{Interventional generation}: guide the noisy intermediate representations by the gradients, or a projection thereof, from the target model \begin{math}f\end{math} s.t. the counterfactual \begin{math}x^{\textrm{CF}}\end{math} elicits a desired output \begin{math}y^{\mathrm{CF}}\end{math} from \begin{math}f\end{math}.
\end{enumerate}
Since intermediate representations of the diffusion model have high-noise levels and may result in mere adversarial perturbations, previous work proposed several schemes to combat these. This included
sharing the encoder between target model and denoising network (Diff-SCM, \citet{sanchez2022diffusion}), albeit at the cost of model-specificity.
Other work (DiME, \citet{jeanneret2022diffusion} \& ACE, \citet{jeanneret2023adversarial}) proposed a computationally intensive iterative approach, where at each diffusion time step, they generated an unconditional, clean image to compute gradients on these unnoisy images, but necessitating backpropagation through the entire diffusion process up to the current time step. This results in computational costs of \begin{math}\mathcal{O}(T^2)\end{math} or \begin{math}\mathcal{O}(T\cdot I)\end{math} for DiME or ACE, respectively, where \begin{math}T\end{math} is the number of diffusion steps and \begin{math}I\end{math} is the number of adversarial attack update steps in ACE.
Lastly, \citet{augustin2022diffusion} (DVCE) introduced a guidance scheme involving a projection between the target model and an auxiliary adversarially robust model. However, this requires the auxiliary model to be trained on a very similar data distribution (and task), which may limit its applicability in settings with restricted data access. Further, we found that counterfactuals generated by DVCE are confounded by the auxiliary model; see \appendixref{sec:dvce_influence} for an extended discussion or Figure 3 of \citet{augustin2022diffusion}.

\begin{comment}
\begin{table}[t]
\centering
\caption{Comparison to previous works using diffusion models for counterfactual explanations. \begin{math}T\end{math} and \begin{math}I\end{math} refer to the diffusion time steps or number of optimization iterations, respectively.
}
\label{tab:comparison}
\resizebox{\linewidth}{!}{
\begin{tabular}{lcccc}
\toprule
Method & Sampling steps & No (auxiliary) robust model & Model-Agnostic & Dataset-Agnostic\\
\midrule
Diff-SCM \citep{sanchez2022diffusion} & \begin{math}\mathcal{O}(T)\end{math} & - & - & -\\
DiME \citep{jeanneret2022diffusion} & \begin{math}\mathcal{O}(T^2)\end{math} & \checkmark & \checkmark & -\\
DVCE \citep{augustin2022diffusion} & \begin{math}\mathcal{O}(T)\end{math} & \checkmark & - & \checkmark\\
ACE \citep{jeanneret2023adversarial} & \begin{math}\mathcal{O}(I\cdot T)\end{math} & \checkmark & \checkmark & - \\
LDCE (Ours) & \begin{math}\mathcal{O}(T)\end{math} & \checkmark & \checkmark & \checkmark \\
\bottomrule
\end{tabular}
}
\end{table}
\end{comment}

\subsection{Overview}\label{sub:latent}
We propose Latent Diffusion Counterfactual Explanations (LDCE) that addresses above limitations: LDCE is model-agnostic, computationally efficient, and alleviates the need for an auxiliary classifier requiring (data-specific, adversarial) training.
LDCE \emph{harnesses the capabilities of recent class- or text-conditional foundation (latent) diffusion models}, augmented with a novel \emph{consensus guidance mechanism} (\secref{sub:guidance}).
The foundational nature of the text-conditional (latent) diffusion model grants LDCE the versatility to be applied across diverse models, datasets (within reasonable bounds), and learning paradigms, as illustrated in \figref{fig:teaser}. Further, we expedite counterfactual generation by operating diffusion models within a perceptually equivalent, \emph{semantic latent space}, as proposed by \citet{rombach2022high}. This also allows LDCE to ``focus on the important, semantic [instead of unimportant, high-frequency details] of the data" (\citet{rombach2022high}, p. 4).
Moreover, our novel consensus guidance mechanism ensures semantically meaningful changes during the reverse diffusion process by leveraging the implicit classifier \citep{ho2022classifier} of class- or text-conditional foundation diffusion models as a filter.
Lastly, note that LDCE is compatible to and will benefit from future advancements of diffusion models.
\algref{alg:method} provides the implementation outline, which will be described in detail below.

\paragraph{Counterfactual generation in latent space}~
We propose generating counterfactual explanations within a perceptually equivalent, lower-dimensional latent space of an autoencoder \citep{esser2021taming} and rewrite \eqref{eq:cf} as follows:
\begin{equation}\label{eq:cf_new}
    x^{\textrm{CF}}=\mathcal{D}(z')\in\argmin\limits_{z'\in\mathcal{Z}=\{\mathcal{E}(x)|x\in\mathcal{X}\}} \lambda_c \mathcal{L}(f(\mathcal{D}(z')),\;y^{\textrm{CF}})+ \lambda_d d(\mathcal{D}(z'),\;x^{\textrm{F}}) \quad .
\end{equation}
Yet, we found that generating counterfactual explanations directly in the autoencoder's latent space \citep{esser2021taming} using a gradient-based approach only results in imperceptible, adversarial changes. Conversely, employing a diffusion model on the latent space produced semantically more meaningful changes. Further, it allows the diffusion model to ``focus on the important, semantic bits of the data" (\citet{rombach2022high}, p. 4) and the autoencoder's decoder \begin{math}\mathcal{D}\end{math} fills in the unimportant, high-frequency image details. This contrasts prior works that used pixel-space diffusion models for counterfactual generation. We refer to this approach as \emph{LDCE-no consensus}.

\subsection{Consensus guidance mechanism}\label{sub:guidance}
\begin{wrapfigure}[14]{r}{0.5\textwidth}
    \centering
    \includegraphics[width=\linewidth]{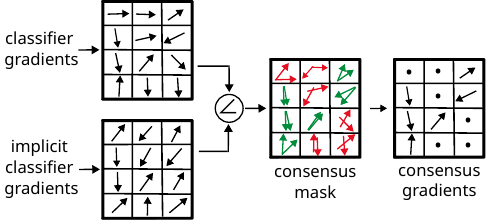}
    \caption{Our proposed consensus guidance mechanism employs a filtering approach of gradients leveraging the implicit classifier of diffusion models as reference for semantic meaningfulness.}
    \label{fig:consensus_guidance}
\end{wrapfigure}
To further mitigate the presence of semantically non-meaningful changes, we introduce a novel consensus guidance mechanism (see \figref{fig:consensus_guidance}). During the reverse diffusion process, our guidance mechanism exclusively allows for gradients from the target model that align with the freely available implicit classifier of a class- or text-conditional diffusion model (\cf, \eqref{eq:implicit}). Consequently, we can use target models out-of-the-box and eliminate the need for auxiliary models that need to be (adversarially) trained on a similar data distribution (and task).
We denote these variants as  \oursc~for class-conditional or \ourst~for text-conditional diffusion models, respectively.

Our consensus guidance mechanism is inspired by the observation that both the gradient of the target model, and the unconditional and conditional score functions of the class- or text-conditional foundation diffusion model (\cf, \eqref{eq:implicit}) estimate \begin{math}\nabla_x \log p(c|x)\end{math}. The main idea of our consensus guidance mechanism is to leverage the latter as a reference for semantic meaningfulness to filter out misaligned gradients of the target model that are likely to result in non-meaningful, adversarial modifications.  More specifically, we compute the angles \begin{math}\alpha_i\end{math} between the target model's gradients \begin{math}\nabla_{z_t} \mathcal{L}(f(\hat{x}_0),\;c)\end{math} and the difference of the conditional and unconditional scores \begin{math}\epsilon_{c}-\epsilon_{uc}\end{math} (\cf, \eqref{eq:implicit}) for each non-overlapping patch, indexed by \begin{math}i\end{math}:
\begin{equation}
    \alpha_i = \angle [ (\sqrt{1-\alpha_t}\nabla_{z_t}\mathcal{L}(f(\hat{x}_0),\;c))_i,\;(\epsilon_{c}-\epsilon_{uc})_i] \quad .
\end{equation}
To selectively en- or disable gradients for individual patches, we introduce an angular threshold \begin{math}\gamma\end{math}:
\begin{equation}
    \phi_i(\alpha_i,\;\gamma,\;\sqrt{1-\alpha_t}\nabla_{z_t}\mathcal{L}(f(\hat{x}_0),\;c))_i,\;\mathbf{o})=\left\{\begin{array}{ll} \sqrt{1-\alpha_t}\nabla_{z_t}\mathcal{L}(f(\hat{x}_0),\;c))_i, & \alpha_i\leq\gamma \\
         \mathbf{o}, & \alpha_i>\gamma\end{array}\right. ,
\end{equation}
where \begin{math}\mathbf{o}\end{math} is the overwrite value (in our case zeros \begin{math}\mathbf{0}\end{math}). Note that by setting the overwrite value \begin{math}\mathbf{o}\end{math} to zeros, only the target model and the unconditional score estimator--which is needed for regularization towards the data manifold--directly influence counterfactual generation. 
\section{Experiments}
\begin{figure}
    \centering
    \begin{subfigure}[b]{0.49\linewidth}
        \includegraphics[width=\linewidth]{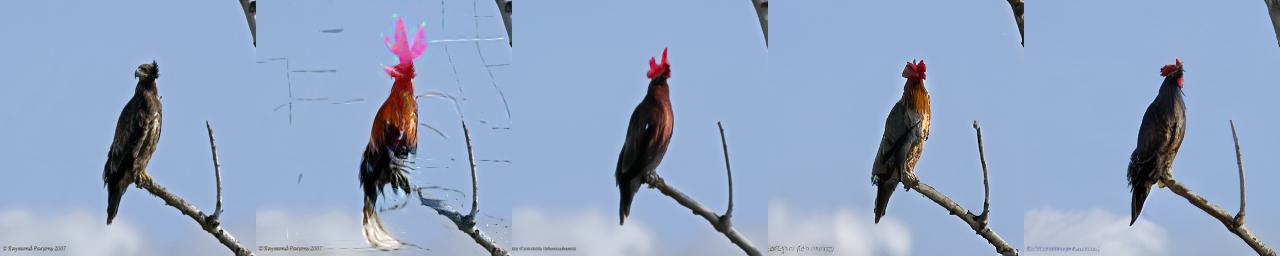}
        \caption{bald eagle $\rightarrow$ rooster}
    \end{subfigure}\hfill
    \begin{subfigure}[b]{0.49\linewidth}
        \includegraphics[width=\linewidth]{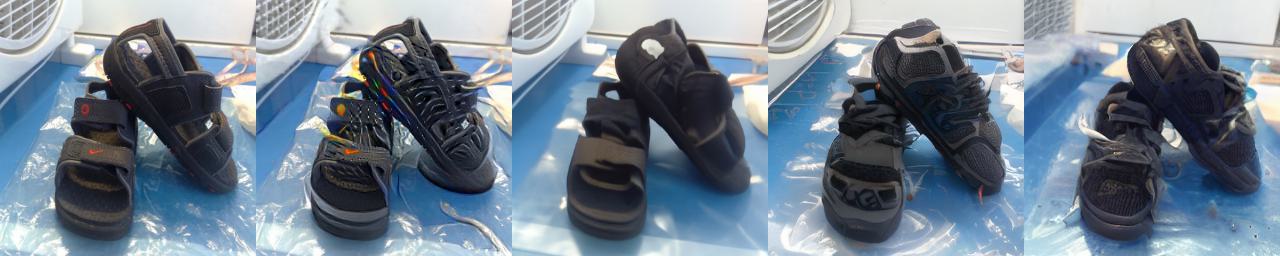}
        \caption{sandal $\rightarrow$ running shoe}
    \end{subfigure}
    \begin{subfigure}[b]{0.49\linewidth}
        \includegraphics[width=\linewidth]{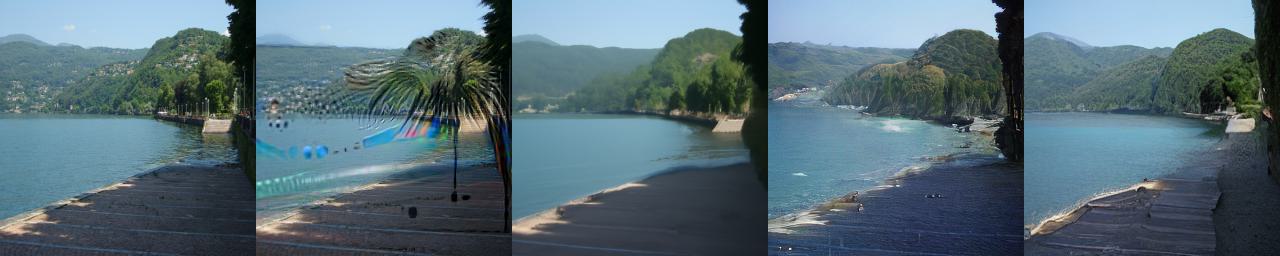}
        \caption{lakeside $\rightarrow$ seashore\label{fig:imagenet_c}}
    \end{subfigure}\hfill
    \begin{subfigure}[b]{0.49\linewidth}
        \includegraphics[width=\linewidth]{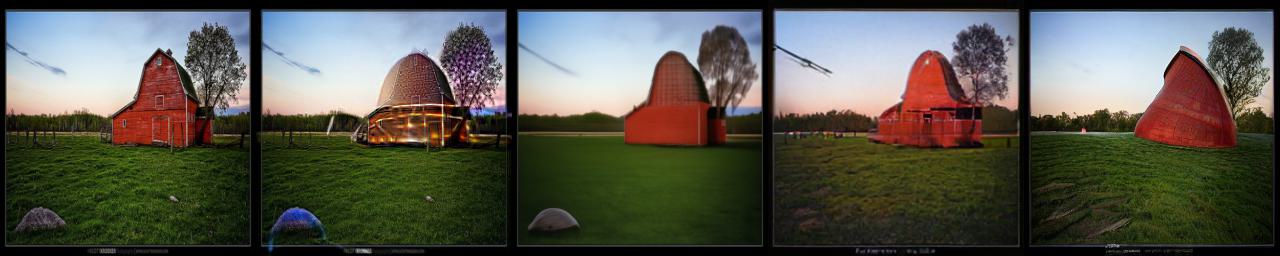}
        \caption{barn $\rightarrow$ planetarium\label{fig:imagenet_d}}
    \end{subfigure}
    \begin{subfigure}[b]{0.49\linewidth}
        \includegraphics[width=\linewidth]{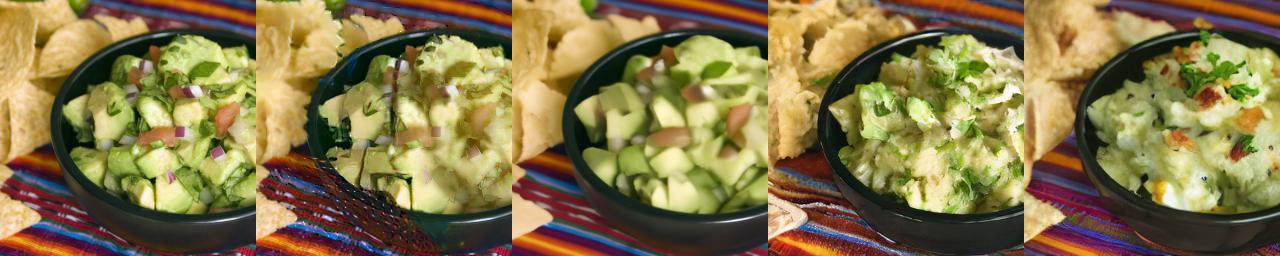}
        \caption{guacamole $\rightarrow$ mashed potato\label{fig:imagenet_e}}
    \end{subfigure}\hfill
    \begin{subfigure}[b]{0.49\linewidth}
        \includegraphics[width=\linewidth]{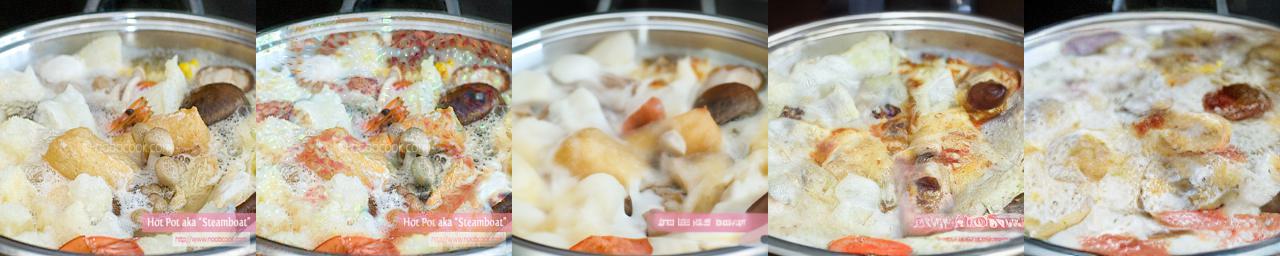}
        \caption{hot pot $\rightarrow$ pizza}
    \end{subfigure}
    \begin{subfigure}[b]{0.49\linewidth}
        \includegraphics[width=\linewidth]{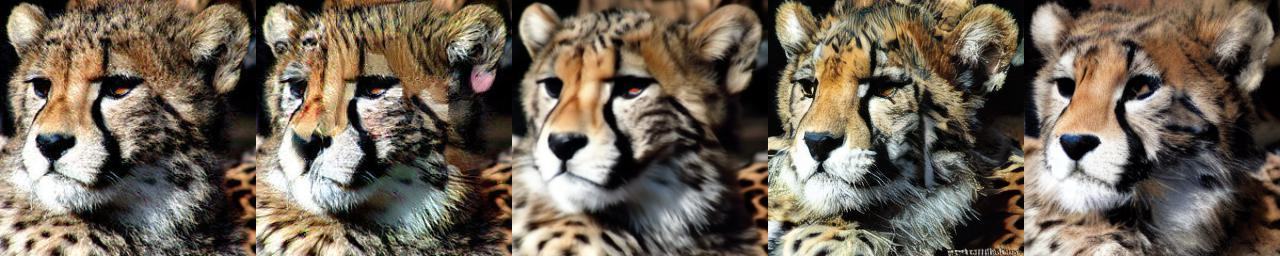}
        \caption{cheetah $\rightarrow$ tiger}
    \end{subfigure}\hfill
    \begin{subfigure}[b]{0.49\linewidth}
        \includegraphics[width=\linewidth]{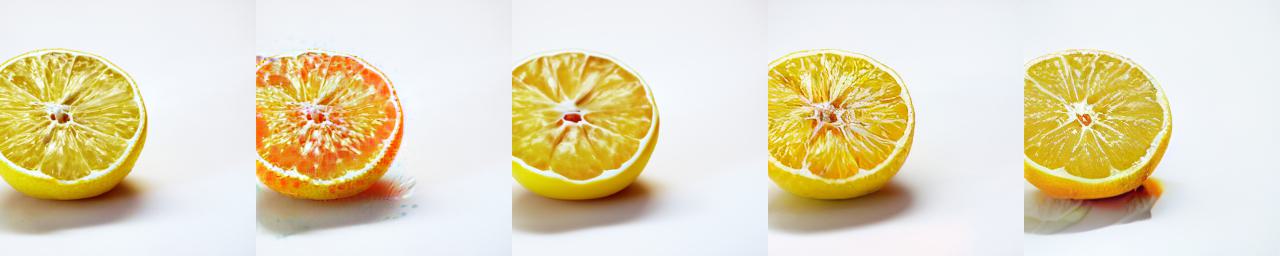}
        \caption{lemon $\rightarrow$ orange}
    \end{subfigure}
    \begin{subfigure}[b]{0.49\linewidth}
        \includegraphics[width=\linewidth]{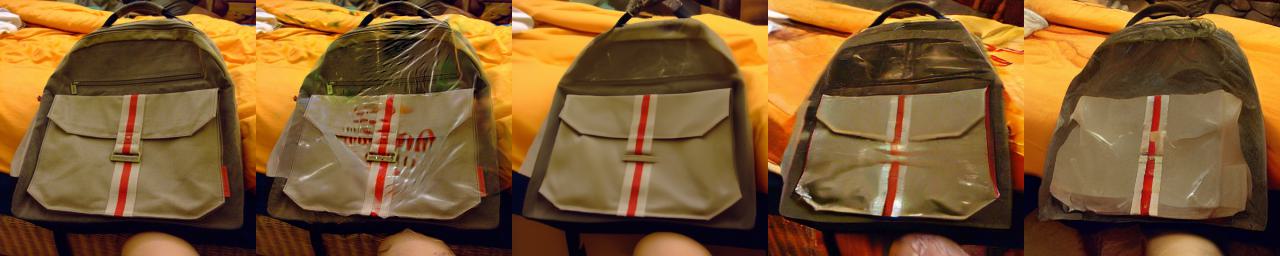}
        \caption{backpack $\rightarrow$ plastic bag}
    \end{subfigure}\hfill
    \begin{subfigure}[b]{0.49\linewidth}
        \includegraphics[width=\linewidth]{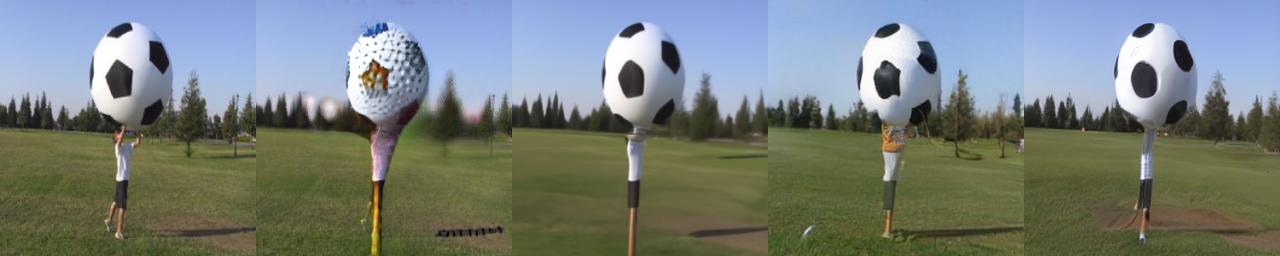}
        \caption{soccer ball $\rightarrow$ golf ball\label{fig:golf_example}}
    \end{subfigure}
    \caption{Qualitative comparison on ImageNet with ResNet-50. Left to right: original image, counterfactual images for SVCE, DVCE, \oursc, and \ourst. \appendixref{sec:more_examples} provides more examples.}
    \label{fig:imagenet}
\end{figure}
\paragraph{Datasets \& models}
We evaluated (and compared) LDCE on ImageNet \citep{deng2009imagenet} (on a subset of 10k images), CelebA HQ \citep{lee2020maskgan}, Oxford Flowers 102 \citep{nilsback2008automated}, and Oxford Pets \citep{parkhi2012cats}. All datasets have image resolutions of 256x256. We used ResNet-50 \citep{he2016deep}, DenseNet-121 \citep{huang2017densely}, OpenCLIP-VIT-B/32 \citep{cherti2022reproducible}, and (frozen) DINO-VIT-S/8 with linear classifier \citep{caron2021emerging} for ImageNet, CelebA HQ, Oxford Pets or Flowers 102, respectively, as target models.
We provide dataset and model licenses in \appendixref{sec:licenses} and further model details in \appendixref{sec:finetuning}.

\paragraph{Evaluation protocol}
We use two protocols for counterfactual target class selection: \textbf{(a) Semantic Hierarchy:} we randomly sample one of the top-4 closest classes based on the shortest path based on WordNet \citep{miller1995wordnet}.
\textbf{(b) Representational Similarity:} we compute the instance-wise cosine similarity with SimSiam  \citep{chen2021exploring} features of the (f)actual images \begin{math}x^{\mathrm{F}}\end{math} and randomly sample one of the top-5 classes. Note that the latter procedure does not require any domain expertise compared to the former.
We adopted the former for ImageNet and the latter for Oxford Pets and Flowers 102. For CelebA HQ, we selected the opposite binary target class.

The evaluation of (visual) counterfactual explanations is inherently challenging: what makes a good counterfactual is arguably very subjective. Despite this, we used various quantitative evaluation criteria covering commonly acknowledged desiderata.
\textbf{(a) Validity:} We used Flip Ratio (FR), \ie, does the generated counterfactual \begin{math}x^{\mathrm{CF}}\end{math} yield the desired output \begin{math}y^{\mathrm{CF}}\end{math}, and COUT \citep{khorram2022cycle} that additionally takes the sparsity of the changes into account. Further, we used the \begin{math}S^3\end{math} criterion \citep{jeanneret2023adversarial} that computes cosine similarity between the (f)actual \begin{math}x^{\textrm{F}}\end{math} and counterfactual \begin{math}x^{\textrm{CF}}\end{math}. While it has been originally introduced a closeness criterion, we found that \begin{math}S^3\end{math} correlates strongly with FR, \ie, we found a high Spearman rank correlation of \begin{math}-0.83\end{math} using the numbers from \tabref{tab:imagenet_ace}.
\textbf{(b) Closeness:} We used L1 and L2 norms to assess closeness. However, note that L$p$ norms can be confounded by unimportant, high-level image details.
\textbf{(c) Realism:} We used FID and sFID \citep{jeanneret2023adversarial} to assess realism.  In contrast to FID, sFID removes the bias caused by the closeness desiderata.
\appendixref{sec:criteria} provides more details.

\paragraph{Implementation details}
We based \oursc~on a class-conditional latent diffusion model trained on ImageNet \citep{rombach2022high} and \ourst~on a fine-tuned variant of Stable Diffusion v1.4 for 256x256 images \citep{pinkey2023mini}. Model licenses and links to the weights are provided in \appendixref{sec:licenses}. For text conditioning, we mapped counterfactual target classes to CLIP-style text prompts \citep{radford2021learning}.
For our consensus guidance scheme, we used spatial regions of size 1x1 and chose zeros as overwrite values. We used L1 as a distance function \begin{math}d\end{math} to promote sparse changes. 
We used a diffusion respacing factor of 2 to expedite counterfactual generation at the cost of a reduction in image quality. We set the weighting factor $\eta$ to 2.
We optimized other hyperparameters (diffusion steps \begin{math}T\end{math} and the other weighing factors
\begin{math}\lambda_c,\;\gamma,\;\lambda_d\end{math}) on 10–20 examples. Note that these hyperparameters control the trade-off of the desiderata validity, closeness, and realism. 
\appendixref{sec:hyperparameters} provides the selected hyperparameters.

\subsection{Qualitative evaluation}\label{sub:qualitative}
\newcommand{\cdownar}{}
\newcommand{\cupar}{}
\begin{figure}
    \centering
    \begin{subfigure}[b]{0.24\linewidth}
        \includegraphics[width=\linewidth]{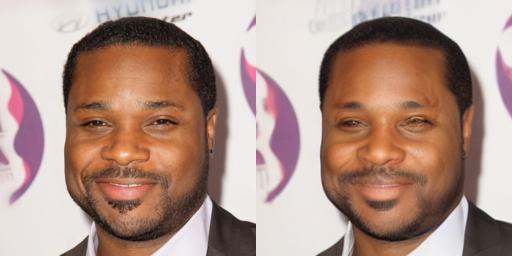}
        \caption{young $\rightarrow$ old}
        \includegraphics[width=\linewidth]{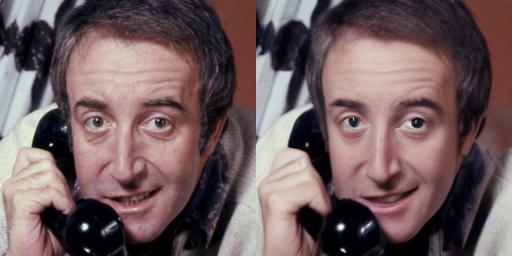}
        \caption{old $\rightarrow$ young\label{fig:celeb_face}}
    \end{subfigure}\hfill
    \begin{subfigure}[b]{0.24\linewidth}
        \includegraphics[width=\linewidth]{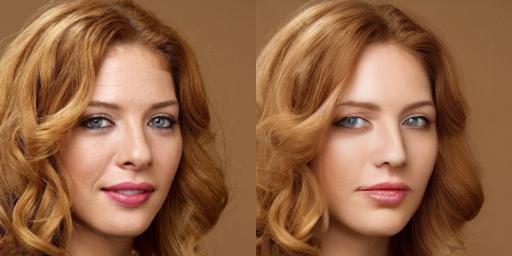}
        \caption{smile $\rightarrow$ no smile}
        \includegraphics[width=\linewidth]{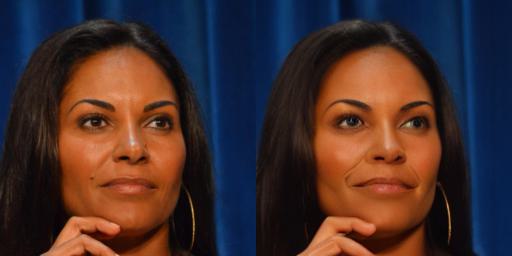}
        \caption{no smile $\rightarrow$ smile\label{fig:celeb_face2}}
    \end{subfigure}\hfill
    \begin{subfigure}[b]{0.24\linewidth}
        \includegraphics[width=\linewidth]{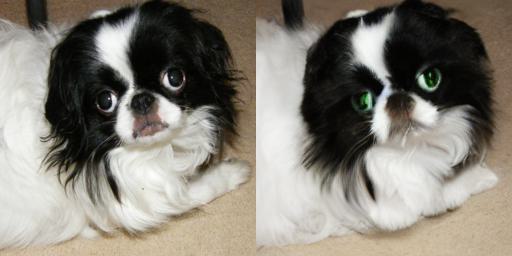}
        \caption{jpn. chin $\rightarrow$ persian cat}
        \includegraphics[width=\linewidth]{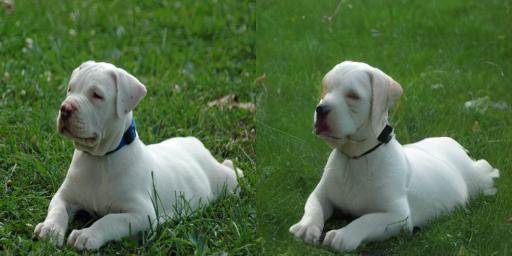}
        \caption{am. bulldog $\rightarrow$ beagle}
    \end{subfigure}\hfill
    \begin{subfigure}[b]{0.24\linewidth}
        \includegraphics[width=\linewidth]{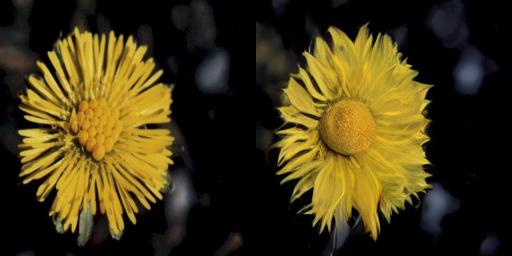}
        \caption{coltsfoot $\rightarrow$ sunflower}
        \includegraphics[width=\linewidth]{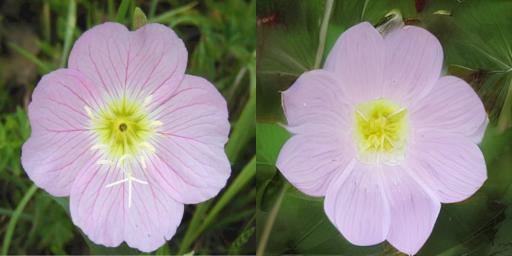}
        \caption{pink primrose $\rightarrow$ lotus}
    \end{subfigure}
    \caption{Qualitative results on CelebA HQ, Oxford Pets, and Flowers 102 using DenseNet-121, CLIP, and DINO with linear classifier, respectively. Left: original image. Right: counterfactual image generated by \ourst. Additional examples are provided in \appendixref{sec:more_examples}.}
    \label{fig:other_datasets}
\end{figure}
Figures \ref{fig:teaser}, \ref{fig:imagenet}, and \ref{fig:other_datasets} show qualitative results for both \oursc~and \ourst~across a diverse range of models (from convolutional networks to transformers) trained on various real-world datasets (from ImageNet to CelebA-HQ, Oxford-Pets, or Flowers-102) with distinct learning paradigms (from supervision, to vision-only or vision-language self-supervision). We observe that \ourst~can introduce local changes, \eg, see \figref{fig:teaser_celeb}, as well as global modifications, \eg, see \figref{fig:teaser_imagenet}. We observe similar local as well as global changes for \oursc~in \figref{fig:imagenet}. Notably, \ourst~can also introduce intricate changes in the geometry of flower petals without being explicitly trained on such data, see \figref{fig:teaser_flower} or the rightmost column of \figref{fig:other_datasets}. Further, we found that counterfactual generation gradually evolves from coarse (low-frequency) features (\eg, blobs or shapes) at the earlier time steps towards more intricate (high-frequency) details (\eg, textures) at later time steps. We explores this further in \appendixref{sec:evolution}. Lastly, both LDCE variants can generate a diverse set of counterfactuals, instead of only single instances, by introducing stochasticity in the abduction step (see \appendixref{sec:diversity} for examples).

\figref{fig:imagenet} also compares LDCE with previous works: SVCE \citep{boreiko2022sparse} and DVCE \citep{augustin2022diffusion}. We found that SVCE often generates high-frequency (\figref{fig:imagenet_c}) or copy-paste-like artifacts (\figref{fig:imagenet_e}). Further, DVCE tends to generate blurry and lower-quality images. This is also reflected in its worse (s)FID scores in \tabref{tab:imagenet_full}. Moreover, note that counterfactuals generated by DVCE are confounded by its auxiliary model; refer to \appendixref{sec:dvce_influence} for an extended discussion.
In contrast to SVCE and DVCE, both LDCE variants generate fewer artifacts, less blurry, and higher-quality counterfactual explanations. However, we also observed failure modes (\eg, distorted secondary objects) and provide examples in \appendixref{sec:failures}. We suspect that some of these limitations are inherited from the underlying foundation diffusion model and, in part, to domain shift. We further discuss these challenges in \Secref{sec:limitations}.

\subsection{Quantitative evaluation}\label{sub:quantitative}
\begin{table}[t]
    \centering
    \caption{Comparison to SVCE and DVCE on ImageNet using ResNet-50.}
    \label{tab:imagenet_full}
    \centering
    \begin{tabular}{lccccc}
    \toprule
        Method & L1 (\begin{math}\downarrow\end{math}) & L2 (\begin{math}\downarrow\end{math}) & FID (\begin{math}\downarrow\end{math}) & sFID (\begin{math}\downarrow\end{math}) & FR (\begin{math}\uparrow\end{math}) \\
        \midrule
        \begin{math}\ell_{1.5}\end{math}-SVCE\begin{math}^{\dag}\end{math} \citep{boreiko2022sparse} & \textbf{5038} & \textit{25} & 22.44 & 28.44 & 83.82 \\ 
        DVCE \citep{augustin2022diffusion} & \textit{6453} & \textbf{24} & 23.94 & 28.99 & 84.0\\
        \midrule
        LDCE-no consensus & 12337 & 41 & 21.70 & \textit{27.10} & \textbf{98.4} \\
        \oursc & 12375 & 42 &  \textbf{14.03} & \textbf{19.25} & 83.1 \\
        \ourst\begin{math}^{\ast}\end{math} & 11577 & 41 & \textit{21.0}\textcolor{white}{0} & \textit{26.5}\textcolor{white}{0} & \textit{84.4}\\
        \bottomrule
        \multicolumn{6}{c}{\begin{math}^\dag\end{math}: used an adversarially robust ResNet-50. \begin{math}^\ast\end{math}: diffusion model not trained on ImageNet.}
    \end{tabular}
\end{table}
We quantitatively compared LDCE to previous work (\begin{math}\ell_{1.5}\end{math}-SVCE \citep{boreiko2022sparse}, DVCE \citep{augustin2022diffusion}, and ACE \citep{jeanneret2023adversarial}) on ImageNet. Note that other previous work is hardly applicable to ImageNet or code is not provided, \eg, C3LT \citep{khorram2022cycle}. Note that we used a multiple-norm robust ResNet-50 \citep{croce2021adversarial} for \begin{math}\ell_{1.5}\end{math}-SVCE since it is tailored for adversarially robust models. Further, we limited our comparison to ACE to their smaller evaluation protocol for ImageNet due to its computationally intensive nature.

\begin{wraptable}[23]{r}{6.5cm}
    \caption{Comparison of \oursc~and \ourst~(\begin{math}^*\end{math}diffusion model not trained on ImageNet) to ACE on ImageNet with ResNet-50.}
    \label{tab:imagenet_ace}
    \resizebox{\linewidth}{!}{
    \footnotesize
    \setlength{\tabcolsep}{3pt}
    \begin{tabular}{c|ccccc} \toprule
        Method        & FID\cdownar  & sFID \cdownar  & S\begin{math}^3\end{math} \cdownar  & COUT \cupar & FR \cupar \\ \midrule
        \multicolumn{6}{c}{\textbf{Zebra -- Sorrel}} \\\midrule
        ACE \begin{math}\ell_1\end{math}  & 84.5 & 122.7 & 0.92 & -0.45 & 47.0\\
        ACE \begin{math}\ell_2\end{math}  & \textbf{67.7} & \textbf{98.4}  & 0.90 & -0.25 & 81.0\\
        \oursc & 84.2 & 107.2  & 0.78 & \textbf{-0.06} & \textbf{88.0} \\
        \ourst\begin{math}^{\ast}\end{math} & 82.4 & 107.2  &  \textbf{0.7113} & -0.2097 & 81.0 \\
        \midrule
        \multicolumn{6}{c}{\textbf{Cheetah -- Cougar}} \\\midrule
        ACE \begin{math}\ell_1\end{math}  & \textbf{70.2}  &100.5 & 0.91 & 0.02   & 77.0 \\
        ACE \begin{math}\ell_2\end{math}  & 74.1 &102.5 & 0.88 & 0.12   & 95.0 \\
        \oursc  & 71.0 & \textbf{91.8} & 0.62 &  \textbf{0.51} & \textbf{100.0} \\
        \ourst\begin{math}^{\ast}\end{math} &  91.2 & 117.0 &  \textbf{0.59} & 0.34 & 98.0 \\
        \midrule
        \multicolumn{6}{c}{\textbf{Egyptian Cat -- Persian Cat}} \\\midrule
        ACE \begin{math}\ell_1\end{math}  & \textbf{93.6} & 156.7& 0.85 & 0.25   & 85.0 \\
        ACE \begin{math}\ell_2\end{math}  & 107.3& 160.4& 0.78 & 0.34   & 97.0 \\
        \oursc & 102.7 & \textbf{140.7}  & 0.63 & 0.52 &  \textbf{99.0}\\
        \ourst\begin{math}^{\ast}\end{math}  & 121.7 & 162.4 &  \textbf{0.61} &  \textbf{0.56} & \textbf{99.0} \\
        \bottomrule
    \end{tabular}
    }
\end{wraptable}
\tabref{tab:imagenet_full} shows that both LDCE variants achieve strong performance for the validity (high FR) and realism (low FID figures) desiderata. Unsurprisingly, we find that SVCE generates counterfactuals that are closer to the (f)actual image (lower L\begin{math}p\end{math} norms) since it specifically constraints optimization within a \begin{math}\ell_{1.5}\end{math}-ball. We also note that unsurprisingly L\begin{math}p\end{math} norms are higher for both LDCE variants than for the other methods since L\begin{math}p\end{math} are confounded by unimportant, high-frequency image details that are not part of our counterfactual optimization, \ie, the decoder just fills in these details. This is corroborated by our qualitative inspection in \Secref{sub:qualitative}.
Furthermore, we found that both LDCE variants consistently outperform ACE across nearly all evaluation criteria; see \tabref{tab:imagenet_ace}. The only exception is FID, which is unsurprising given that ACE enforces sparse changes, resulting in counterfactuals that remain close to the (f)actual images. This is affirmed by ACE's lower sFID scores, which accounts for this.
We provide quantitative comparisons with other methods on CelebA HQ in \appendixref{sec:celeb_results}. Despite not being trained specifically for faces, \ourst~yielded competitive results to other methods.

Lastly, we evaluated the computational efficiency of diffusion-based counterfactual methods. Specifically, we compared throughput on a single NVIDIA RTX 3090 GPU with \SI{24}{\gigabyte} memory for 20 batches with maximal batch size. We report that \ourst~only required \SI{156}{\second} for four counterfactuals, whereas DVCE needed \SI{210}{\second} for four counterfactuals, and ACE took \SI{184}{\second} for just a single counterfactual. Note that DiME, by design, is even slower than ACE. Above throughput differences translate to substantial speed-ups of \SI{34}{\percent} and \SI{371}{\percent} compared to DVCE or ACE, respectively.

Above results highlight LDCE as a strong counterfactual generation method. In particular, \ourst~achieves on par and often superior performance compared to previous approaches, despite being the only method not requiring any component to be trained on the same data as the target model--a property that, to the best of our knowledge, has not been available in any previous work and also enables applicability in real-world scenarios where data access may be restricted.

\subsection{Identification and resolution of model errors}
\begin{figure}
    \begin{subfigure}[b]{0.33\linewidth}
        \includegraphics[width=\linewidth]{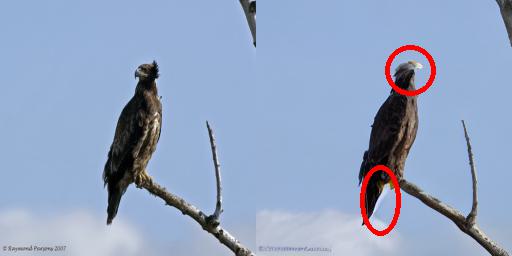}
        \caption{kite $\rightarrow$ bald eagle}
    \end{subfigure}\hfill
    \begin{subfigure}[b]{0.33\linewidth}
        \includegraphics[width=\linewidth]{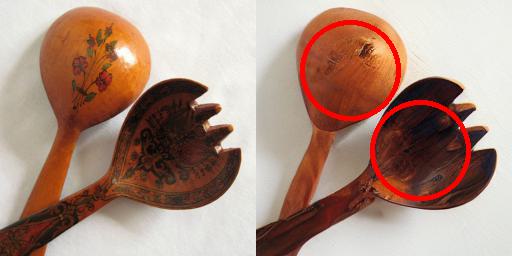}
        \caption{maraca $\rightarrow$ wooden spoon}
    \end{subfigure}\hfill
    \begin{subfigure}[b]{0.33\linewidth}
        \includegraphics[width=\linewidth]{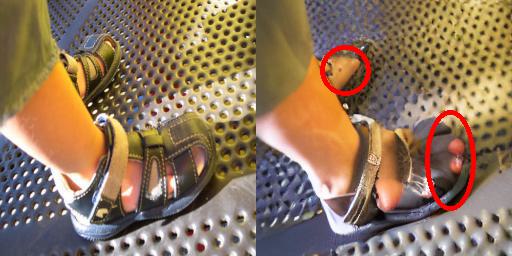}
        \caption{running shoe $\rightarrow$ sandal}
    \end{subfigure}
    \caption{Examples of identified classification errors of ResNet-50. Left: original image that is misclassified. Right: counterfactual that is correctly classified. The red ellipses were added manually.}
    \label{fig:model_errors}
\end{figure}
Counterfactual explanations should not solely generate high-quality images, like standard image generation, editing or prompt-to-prompt methods \citep{hertz2023prompttoprompt}, but should serve to better understand model behavior. More specifically, we showcase how \ourst~can effectively enhance our model understanding of ResNet-50 trained with supervision on ImageNet in the context of misclassifications. To this end, we generated counterfactual explanations of a ResNet-50's misclassifications towards the true class. As illustrated in \figref{fig:model_errors}, this elucidates missing or misleading features in the original image that lead to a misclassification. For instance, it reveals that ResNet-50 may misclassify young bald eagles primarily due to the absence of their distinctive white heads (and tails), which have yet to fully develop. Similarly, we found that painted wooden spoons may be misclassified as maraca, while closed-toe sandals may be confused with running shoes.

To confirm that these findings generalize beyond single instances, we first tried to synthesize images for each error type with InstructPix2Pix \citep{brooks2023instructpix2pix}, but found that it often could not follow the instructions; indicating that these model errors transcend ResNet-50 trained with supervision. 
Thus, we searched for 50 images on the internet, and report classification error rates of \SI{88}{\percent}, \SI{74}{\percent}, and \SI{48}{\percent} for the bald eagle, wooden spoon, or sandal model errors, respectively.
Finally, we used these images to finetune the last linear layer of ResNet-50. To this end, we separated the 50 images into equally-sized train and test splits; finetuning details are provided in \appendixref{sec:finetuning_errors}. This  effectively mitigated these model errors and reduced error rates on the test set by \SI{40}{\percent}, \SI{32}{\percent}, or \SI{16}{\percent}, respectively.

\section{Limitations}\label{sec:limitations}
The main limitation of LDCE is its slow counterfactual generation, which hinders real-time, interactive applications. However, advancements in distilling diffusion models \citep{salimans2022progressive,song2023consistency,meng2023distillation} or speed-up techniques \citep{dao2022flashattention,bolya2023token} offer promise in mitigating this limitation.
Another limitation is the requirement for hyperparameter optimization. Although this is very swift, it still is necessary due to dataset differences and diverse needs in use cases.
Lastly, while contemporary foundation diffusion models have expanded their data coverage \citep{schuhmann2021laion}, they may not perform as effectively in specialized domains, \eg, for biomedical data, or may contain (social) biases \citep{bianchi2023easily,luccioni2023stable}.

\section{Conclusion}
We introduced LDCE to generate semantically meaningful counterfactual explanations using class- or text-conditional foundation (latent) diffusion models, combined with a novel consensus guidance mechanism. We show LDCE's  versatility across diverse models learned with diverse learning paradigms on diverse datasets, and demonstrate its applicability to better understand model errors and resolve them. Future work could employ our consensus guidance mechanism beyond counterfactual generation and incorporate spatial or textual priors.

\subsection*{Broader impact}
Counterfactual explanations aid in understanding model behavior, can reveal model biases, \etc. By incorporating latent diffusion models \citep{rombach2022high}, we make a step forward in reducing computational demands in the generation of counterfactual explanations.
However, counterfactual explanations may be manipulated \citep{slack2021counterfactual} or abused. Further, (social) biases in the foundation diffusion models \citep{bianchi2023easily,luccioni2023stable} may also be reflected in counterfactual explanations, resulting in misleading explanations.

\ifboolexpr{not togl {iclrsubmission}}{
\subsubsection*{Author contributions}
\small
Project idea: S.S.;
project lead: S.S. \& K.F.;
conceptualization of consensus guidance mechanism: K.F. with input from S.S.;
method implementation: K.F. \& S.S.;
hyperparameter optimization: K.F.;
implementation and execution of experiments: S.S. \& K.F. with input from M.A. \& T.B.;
visualization: S.S. with input from K.F. (experimental results) \& M.A. with input from S.S. \& K.F. (\figref{fig:consensus_guidance});
interpretation of findings: K.F. \& S.S. with input from M.A. \& T.B.;
guidance \& feedback: M.A. \& T.B.;
funding acquisition: T.B.;
paper writing: S.S. \& K.F. crafted the first draft and all authors contributed to the final version.
\normalsize

\subsubsection*{Acknowledgments}
\small
This work was funded by the Bundesministerium für Umwelt, Naturschutz, nukleare Sicherheit und Verbraucherschutz (BMUV, German Federal Ministry for the Environment, Nature Conservation, Nuclear Safety and Consumer Protection) based on a resolution of the German Bundestag (67KI2029A) and the Deutsche Forschungsgemeinschaft (DFG, German Research Foundation) under grant number 417962828.
K.F. acknowledges supported by the Deutscher Akademischer Austauschdienst (DAAD, German Academic Exchange Service) as part of the ELIZA program.
\normalsize
}

\bibliography{egbib}
\bibliographystyle{iclr2024_conference}

\appendix
\section{Influence of the adversarial robust model in DVCE}\label{sec:dvce_influence}
\begin{wrapfigure}[18]{r}{6cm}
    \centering
        \includegraphics[width=\linewidth]{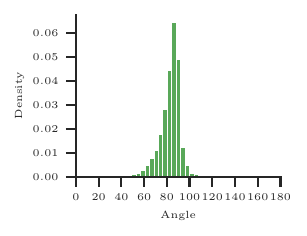}
        \caption{Distribution of angles between the gradient of the target and the robust classifier over 100 images over 100 timesteps.}
     \label{fig:histo_gradients}
\end{wrapfigure}
\begin{figure}[t]
    \centering
    \begin{subfigure}[b]{0.49\linewidth}
        \includegraphics[width=\linewidth]{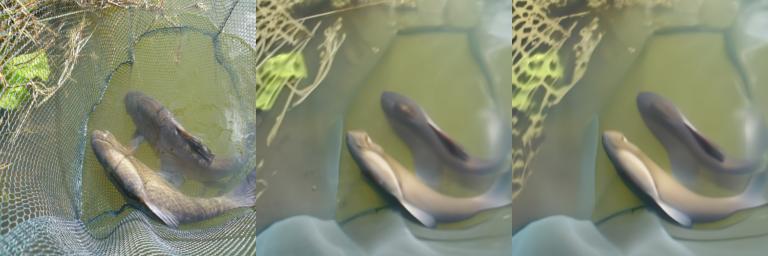}
        \caption{tench $\rightarrow$ eel}
    \end{subfigure}
    \begin{subfigure}[b]{0.49\linewidth}
        \includegraphics[width=\linewidth]{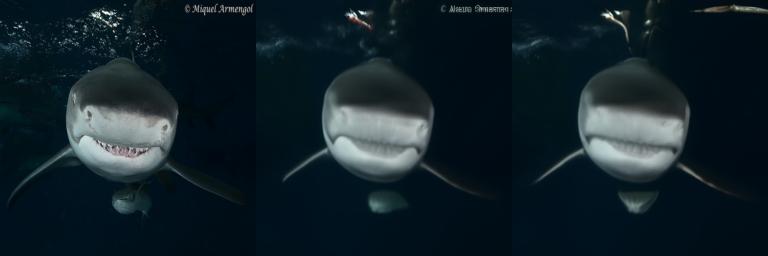}
        \caption{tiger shark $\rightarrow$ great white shark}
    \end{subfigure}
    \begin{subfigure}[b]{0.49\linewidth}
        \includegraphics[width=\linewidth]{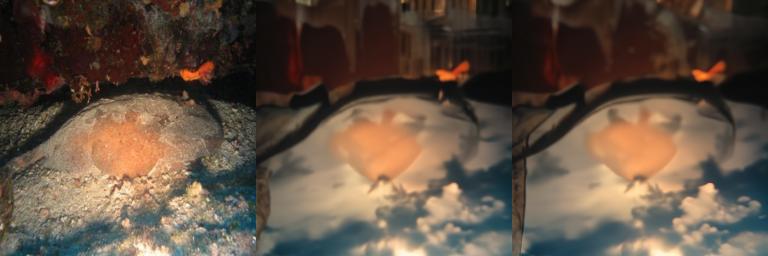}
        \caption{electric ray $\rightarrow$ great white shark}
    \end{subfigure}
    \begin{subfigure}[b]{0.49\linewidth}
        \includegraphics[width=\linewidth]{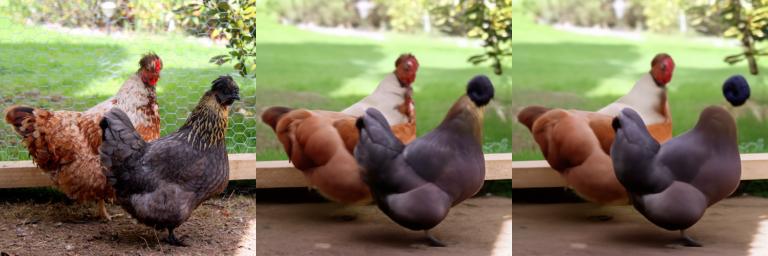}
        \caption{hen $\rightarrow$ rooster}
    \end{subfigure}
    \begin{subfigure}[b]{0.49\linewidth}
        \includegraphics[width=\linewidth]{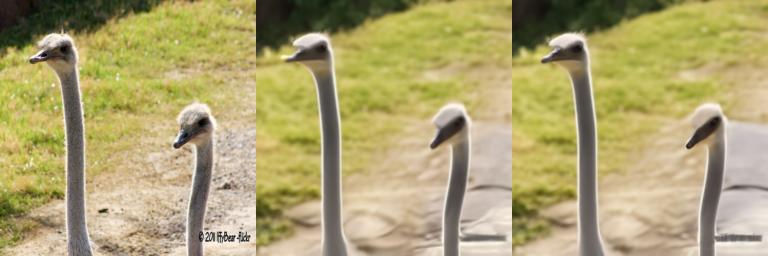}
        \caption{ostrich $\rightarrow$ african grey}
    \end{subfigure}
    \begin{subfigure}[b]{0.49\linewidth}
        \includegraphics[width=\linewidth]{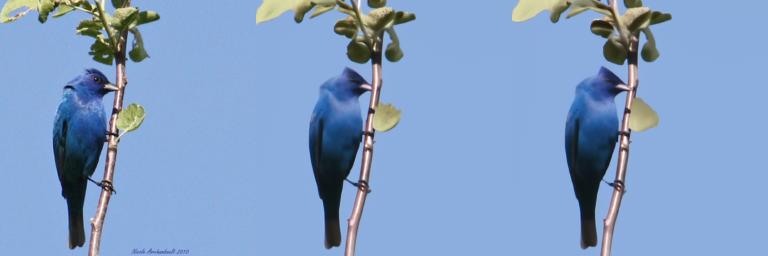}
        \caption{indigo bunting $\rightarrow$ goldfinch}
    \end{subfigure}

    \caption{Qualitative examples illustrating the marginal influence of the target model in DVCE. From left to right: original image, counterfactual images generated using DVCE with cone projection using an angular threshold of $30^\circ$ and the robust classifier, and DVCE using the robust classifier only, \ie, without the target model.}
    \label{fig:robustness_examples}
\end{figure}

DVCE \citep{augustin2022diffusion} uses a projection technique to promote semantically meaningful changes, \ie, they project the unit gradient of an adversarially robust model onto a cone around the unit gradient of the target model, when the gradient directions disagree, \ie, the angle between them exceeds an angular threshold \begin{math}\alpha\end{math}. More specifically, \citet{augustin2022diffusion} defined their cone projection as follows:
\begin{equation}
    P_{\textrm{cone}(\alpha,v)}[w] :=\left\{\begin{array}{ll} \langle u,w\rangle u, & \angle(w,v) >\alpha \\
         v, & \textrm{else}\end{array}\right. ,
\end{equation}
where \begin{math}v,w\end{math} are the unit gradients of the target classifier and adversarially robust classifier, respectively, \begin{math}\textrm{cone}(\alpha,v):=\{w\in\mathbb{R}^d:\angle(v,w)\leq\alpha\}\end{math}, and 
\begin{equation}
    u=\sin{\alpha}\frac{P_{v^\perp}(w)}{||P_{v^\perp}(w)||_2}+\cos{\alpha}\frac{v}{||v||_2} \qquad ,
\end{equation}
where \begin{math}P_{v^\perp(w)}:=w-\frac{\langle w,v\rangle}{\langle v,v\rangle}v\end{math}.
However, due to the high dimensionality of the unit gradients (\begin{math}\mathbb{R}^{256\cdot 256\cdot 3}\end{math}), they are nearly orthogonal with high probability.\footnote{Note that two randomly uniform unit vectors a nearly orthogonal with high probability in high-dimensional spaces. This can be proven via the law of large numbers and central limit theorem.}
In fact, we empirically observed that ca. \SI{97.73}{\percent} of the gradients pairs have angles larger than 60\begin{math}^{\circ}\end{math} during the counterfactual generation of 100 images from various classes; see \figref{fig:histo_gradients}. 
As a result, we almost always use the cone projection and found that the counterfactuals are substantially influenced by the adversarially robust model:
Figure 3 of \citet{augustin2022diffusion} and \figref{fig:robustness_examples} highlight that the target model has a limited effect in shaping the counterfactuals.
Consequently, we \emph{cannot} attribute the changes of counterfactual explanations solely to the target model since they are confounded by the auxiliary adversarially robust model.

\section{Dataset and model licenses}\label{sec:licenses}
\tabref{tab:licenses_datasets} and \ref{tab:licenses_models} provide licenses and URLs of the datasets or models, respectively, used in our work.
Our implementation is built upon \citet{rombach2022high} (License: Open RAIL-M, URL: \url{https://github.com/CompVis/stable-diffusion}) and provided at \codeURL~(License: MIT).
\begin{table}[h]
    \centering
    \caption{Licenses and URLs for the datasets used in our experiments.}
    \label{tab:licenses_datasets}
    \resizebox{\linewidth}{!}{
    \begin{tabular}{lcc}
        \toprule
        Dataset & License & URL\\\midrule
        CelebAMask-HQ \citep{lee2020maskgan} & CC BY 4.0 & \url{https://github.com/switchablenorms/CelebAMask-HQ} \\
        Oxford Flowers 102 \citep{nilsback2008automated} & GNU & \url{https://www.robots.ox.ac.uk/~vgg/data/flowers/102/} \\
        ImageNet \citep{deng2009imagenet} & Custom &  \url{https://www.image-net.org/index.php} \\
        Oxford Pet \citep{parkhi2012cats} & CC BY-SA 4.0 & \url{https://www.robots.ox.ac.uk/~vgg/data/pets/} \\\bottomrule
    \end{tabular}
    }
\end{table}
\begin{table}[h]
    \centering
    \caption{Licenses and URLs for the target and diffusion models used in our experiments.}
    \label{tab:licenses_models}
    \resizebox{\linewidth}{!}{
    \begin{tabular}{lcc}
        \toprule
        Models & License & URL\\\midrule
        \begin{tabular}{@{}l@{}}ImageNet class- \\ conditional LDM \citep{rombach2022high}\end{tabular} & MIT  & \url{https://github.com/CompVis/latent-diffusion} \\
         \begin{tabular}{@{}l@{}}Mini Stable \\ diffusion 1.4 \citep{pinkey2023mini}\end{tabular} &  \begin{tabular}{@{}c@{}}CreativeML \\ Open RAIL-M \citep{rombach2022high}\end{tabular} &  \url{https://huggingface.co/justinpinkney/miniSD} \\\midrule
        ResNet-50 for ImageNet \citep{he2016deep,torchvision2016} & BSD 3 & \url{https://github.com/pytorch/vision} \\
        Adv. robust ResNet-50 for ImageNet \citep{boreiko2022sparse} & MIT & \url{https://github.com/valentyn1boreiko/SVCEs_code} \\
        DenseNet-121 \citep{huang2017densely} for CelebA HQ \citep{jacob2022steex} & Apache 2 & \url{https://github.com/valeoai/STEEX} \\
        DINO for Oxford Flowers 102 \citep{caron2021emerging} & Apache 2 & \url{https://github.com/facebookresearch/dino} \\
        OpenCLIP for Oxford Pets \citep{cherti2022reproducible} & Custom & \url{https://github.com/mlfoundations/open_clip} \\\midrule
        SimSiam \citep{chen2021exploring} & CC BY-NC 4.0 & \url{https://github.com/facebookresearch/simsiam} \\
        CelebA HQ Oracle \citep{jacob2022steex} & Apache 2 & \url{https://github.com/valeoai/STEEX} \\
        \begin{tabular}{@{}l@{}}Ported VGGFace2 \\model from \citep{cao2018vggface2}\end{tabular} & MIT & \url{https://github.com/cydonia999/VGGFace2-pytorch} \\\bottomrule
    \end{tabular}
    }
\end{table}

\section{Model details}\label{sec:finetuning}
Below, we provide model details:
\begin{itemize}
    \item \textbf{ResNet50 \citep{he2016deep} on ImageNet \citep{deng2009imagenet}}:
    We used the pretrained ResNet-50 model provided by torchvision \citep{torchvision2016}.
    \item \textbf{DenseNet-121 \citep{huang2017densely} on CelebA HQ \citep{lee2020maskgan}}: We used the pretrained DenseNet-121 model provided by \citet{jacob2022steex}.
    \item \textbf{OpenCLIP \citep{radford2021learning,cherti2022reproducible} on Oxford Pets \citep{parkhi2012cats}}: We used the provided weights of OpenCLIP ViT-B/32 \citep{dosovitskiy2021vit}, and achieved a top-1 zero-shot classification accuracy of \SI{90.5}{\percent} using CLIP-style prompts \citep{radford2021learning}.
    \item \textbf{DINO+linear \citep{caron2021emerging} on Oxford Flowers 102 \citep{nilsback2008automated}}: We used the pretrained DINO ViT-S/8 model, added a trainable linear classifier, and trained it on Oxford Flowers 102 for 30 epochs. We used SGD with a learning rate of 0.001 and momentum of 0.9, and cosine annealing \citep{loshchilov2018decoupled}. The model achieved a top-1 classification accuracy of \SI{92.82}{\percent}.
\end{itemize}

\section{Evaluation criteria for counterfactual explanations}\label{sec:criteria}
In this section, we discuss the evaluation criteria used to quantitatively assess the quality of counterfactual explanations. Even though quantitative assessment of counterfactual explanations is arguably very subjective, these evaluation criteria build a basis of quantitative evaluation based on the  commonly recognized desiderata validity, closeness, and realism.

\subsection*{Flip Ratio (FR)}
This criterion focuses on assessing the validity of \begin{math}N\end{math} counterfactual explanations by quantifying the degree to which the original class label \begin{math}y_i^{\textrm{F}}\end{math} of the original image \begin{math}x_i^\textrm{F}\end{math} flips the target classifier's prediction \begin{math}f\end{math} to the counterfactual target class \begin{math}y_i^{\textrm{CF}}\end{math} for the counterfactual image \begin{math}x_i^{\textrm{CF}}\end{math}:
\begin{equation}
\text{FR} = \frac{{\sum\limits_{i=1}^{N} \mathbb{I}(f(x^{\textrm{CF}}_i) = y_i^{\textrm{CF}})}}{{N}} \qquad ,
\end{equation}
where \begin{math}\mathbb{I}\end{math} is the indicator function.

\subsection*{Counterfactual Transition (COUT)}
Counterfactual Transition (COUT) \citep{khorram2022cycle} measures the sparsity of changes in counterfactual explanations, incorporating validity and sparsity aspects. It quantifies the impact of perturbations introduced to the (f)actual image \begin{math}x^\textrm{F}\end{math} using a normalized mask \begin{math}m\end{math} that represents relative changes compared to the counterfactual image \begin{math}x^\textrm{CF}\end{math}, \ie, \begin{math}m=\delta(||x^{\textrm{F}}-x^{\textrm{CF}}||_1)\end{math}, where \begin{math}\delta\end{math} normalizes the absolute difference to \begin{math}[0,1]\end{math}.
We progressively perturb \begin{math}x^\textrm{F}\end{math} by inserting top-ranked pixel batches from \begin{math}x^{\textrm{CF}}\end{math} based on these sorted mask values.

For each perturbation step \begin{math}t \in \{0, \ldots, T\}\end{math}, we record the output scores of the classifier \begin{math}f\end{math} for the (f)actual class \begin{math}y^\textrm{F}\end{math} and the counterfactual class \begin{math}y^\textrm{CF}\end{math} throughout the transition from \begin{math}x^0 =x^{\textrm{F}}\end{math} to \begin{math}x^T =x^{\textrm{CF}}\end{math}. From this, we can compute the COUT score:
\begin{equation}
\textrm{COUT} = \textrm{AUPC}(y^{\textrm{CF}}) - \textrm{AUPC}(y^{\textrm{F}})\in[-1,1] \qquad ,
\end{equation}
where the area under the Perturbation Curve (AUPC) for each class \begin{math}y \in \{y^\textrm{F}, y^\textrm{CF}\}\end{math} is defined as follows:
\begin{equation}
\textrm{AUPC}(y) = \frac{1}{T} \sum_{t=0}^{T-1} \frac{1}{2} \left( f_y\left(x^{(t)}\right) + f_y\left(x^{(t+1)}\right) \right) \in [0,1] \qquad .
\end{equation}
A high COUT score indicates that a counterfactual generation approach finds sparse changes that flip classifiers' output to the counterfactual class.

\subsection*{SimSiam Similarity (S3)}
This criterion measures the cosine similarity between a counterfactual image \begin{math}x^{\textrm{CF}}\end{math} and its corresponding (f)actual image \begin{math}x^{\textrm{F}}\end{math} in the feature space of a self-supervised SimSiam model \begin{math}S\end{math} \citep{chen2021exploring}:
\begin{equation}
{S^3(x^{\textrm{CF}}, x^{\textrm{F}}) = \frac{{ \mathcal{S}(x^{\textrm{CF}}) \cdot  \mathcal{S}(x^{\textrm{F}})}}{{\lVert  \mathcal{S}(x^{\textrm{CF}}) \rVert \lVert \mathcal{S}(x^{\textrm{F}}) \rVert}}}    \quad .
\end{equation}

\subsection*{Lp norms}
\begin{math}Lp\end{math} norms serve as closeness criteria by quantifying the magnitude of the changes between the counterfactual image \begin{math}x_{i}^{\textrm{CF}}\end{math} and original image \begin{math}x_{i}^{\textrm{F}}\end{math}:
\begin{equation}
L_p = \frac{1}{N} \sum_{i=1}^{N} \| d_{i} \|_{p} \qquad ,
\end{equation}
where \begin{math}0<p\leq\infty\end{math} and \begin{math}C,H,W\end{math} are the number of channels, image height, and image width, respectively, and
\begin{equation}
\| d_{i} \|_{p} = \left( \sum_{c=1}^{C} \sum_{h=1}^{H} \sum_{w=1}^{W} |x_{i,c,h,w}^{\textrm{F}} - x_{i,c,h,w}^{\textrm{CF}}|^{p} \right)^{\frac{1}{p}} \qquad .
\end{equation}

Note that \begin{math}Lp\end{math} norms can be confounded by unimportant, high-frequency image details.

\subsection*{Mean Number of Attribute Changes (MNAC)}
Mean Number of Attribute Changes (MNAC) quantifies the average number of attributes modified in the generated counterfactual explanations. 
It uses an oracle model  \begin{math}O_a\end{math} (\ie, VGGFace2 model \citep{cao2018vggface2}) which predicts the probability of each attribute \begin{math}a \in \mathcal{A}\end{math}, where  \begin{math}\mathcal{A}\end{math} is the entire attributes space.
MNAC is defined as follows:
\begin{equation}
\text{MNAC} =\frac{1}{N} \sum_{i=1}^{N} \sum_{a \in \mathcal{A} }^{} \left[ \mathbb{I}\left( \mathbb{I}\left( O_a(x^{\textrm{CF}}_i) > \beta \right) \neq \mathbb{I}\left( O_a(x^{\textrm{F}}_i) > \beta \right) \right) \right] \qquad ,
\end{equation}
where \begin{math}{\beta}\end{math} is a threshold (typically set to 0.5) that determines the presence of attributes.
MNAC quantifies the counterfactual method's changes to the query attribute \begin{math}q\end{math}, while remaining independent of other attributes. However, a higher MNAC value can wrongly assign accountability for spurious correlations to the counterfactual approach, when in fact they may be artifacts of the classifier.

\subsection*{Correlation Difference (CD)}
Correlation Difference (CD) \citep{jeanneret2022diffusion} evaluates the ability of counterfactual methods to identify spurious correlations by comparing the Pearson correlation coefficient \begin{math}c^{q,a}(x)\end{math}, of the relative attribute changes \begin{math}\delta^{q}\end{math} and \begin{math}\delta^{a}\end{math}, before and after applying the counterfactual method. For each attribute \begin{math}a\end{math}, the attribute change \begin{math}\delta^{a}\end{math} is computed between pairs of samples \begin{math}i\end{math} and \begin{math}j\end{math}, as \begin{math}\delta^{a}_{i,j} = \hat{y}^a_i - \hat{y}^a_j\end{math}, using the predicted probabilities \begin{math}\hat{y}^a_i\end{math} and \begin{math}\hat{y}^a_j\end{math} from the oracle model \begin{math}O\end{math} (\ie, VGGFace2 model \citep{cao2018vggface2}).
The CD for a query attribute \begin{math}q\end{math} is then computed as:
\begin{equation}
\textrm{CD}_q = \frac{1}{N} \sum_{i=1}^{N} \sum_{a \in \mathcal{A} }^{}  |c^{q,a}(x^{\textrm{CF}})-c^{q,a}(x^\textrm{F})| \qquad .
\end{equation}

\subsection*{Face Verification Accuracy (FVA) \& Face Similarity (FS)}
Face Verification Accuracy (FVA) quantifies whether counterfactual explanations for face attributes maintain identity while modifying the target attribute using the VGGFace2 model \citep{cao2018vggface2}, or not. Alternatively, \citet{jeanneret2023adversarial} proposed Face Similarity (FS) that addresses thresholding issues and the abrupt transitions in classifier decisions in FVA when comparing the (f)actual image \begin{math}x^{\textrm{F}}\end{math} and its corresponding counterfactual \begin{math}x^{\textrm{CF}}\end{math}. FS directly measures the cosine similarity between the feature encodings, providing a more continuous assessment (similar to S\begin{math}^3\end{math}).

\subsection*{Fréchet Inception Distances (FID \& sFID)}
Fréchet Inception Distance (FID) \citep{heusel2017gans} and split FID (sFID) \citep{jeanneret2023adversarial} evaluate the realism of generated counterfactual images by measuring the distance on the dataset level between the InceptionV3 \citep{szegedy2016rethinking} feature distributions of real and generated images:
\begin{equation}
\begin{aligned}
\text{FID} &= \|\mu_{\textrm{F}} - \mu_{\textrm{CF}}\|_2^2 + \text{Tr}(\Sigma_{\textrm{F}} + \Sigma_{\textrm{CF}} - 2\sqrt{\Sigma_{\textrm{F}}\Sigma_{\textrm{CF}}}) \qquad ,
\end{aligned}
\end{equation}
where \begin{math}\mu_{\textrm{F}},\mu_{\textrm{CF}}\end{math} and \begin{math}\Sigma_{\textrm{F}},\Sigma_{\textrm{CF}}\end{math} are the feature-wise mean or covariance matrices of the InceptionV3 feature distributions of real and generated images, respectively.
However, there is a strong bias in FID towards counterfactual approaches that barely alter the pixels of the (f)actual inputs. To address this, \citet{jeanneret2023adversarial} proposed to split the dataset into two subsets: generate counterfactuals for one subset, compute FID between those counterfactuals and the (f)actual inputs of the untouched subset, and vice versa, and then take the mean of the resulting FIDs.

\section{Hyperparameters}\label{sec:hyperparameters}
\begin{table}[t]
\centering
\caption{Manually-tuned hyperparameters.}
\label{tab:hyperparameters}
\begin{tabular}{lccccc}
    \toprule
     & \oursc & \multicolumn{4}{c}{\ourst}\\
     \cline{3-6}
    Hyperparameters & ImageNet & ImageNet & CelebA HQ & Flowers & Oxford Pets\\
    \midrule
    consensus threshold \begin{math}\gamma\end{math}& 45\begin{math}^{\circ}\end{math} & 50\begin{math}^{\circ}\end{math} & 55\begin{math}^{\circ}\end{math} & 45\begin{math}^{\circ}\end{math} & 45\begin{math}^{\circ}\end{math} \\
    starts  timestep \begin{math}T\end{math} &  191 & 191 & 200& 250 & 191\\
    classifier weighting \begin{math}\lambda_c\end{math} &  2.3 & 3.95 & 4.0 & 3.4 & 4.2\\
    distance weighting \begin{math}\lambda_d\end{math} & 0.3 & 1.2 & 3.3  & 1.2 & 2.4\\\bottomrule
\end{tabular}
\end{table}
\tabref{tab:hyperparameters} provides our manually tuned hyperparameters.
For our \ourst, we transform the counterfactual target classes \begin{math}y^{\textrm{CF}}\end{math} to CLIP-style text prompts \citep{radford2021learning}, as follows:
\begin{itemize}
    \item ImageNet: \texttt{a photo of a \{category name\}.}
    \item CelebA HQ: \texttt{a photo of a \{attribute name\} person.} (\texttt{attribute name}~\begin{math}\in\end{math}~\{non-smiling, smiling, old, young\}).
    \item Oxford Flowers 102: \texttt{a photo of a \{category name\}, a type of flower.}
    \item Oxford Pets: \texttt{a photo of a \{category name\}, a type of pet.}
\end{itemize}
We note that more engineered prompts may yield better counterfactual explanations, but we leave such studies for future work.

\section{Changes over the course of the counterfactual generation}\label{sec:evolution}
\begin{figure}
    \centering
    \begin{subfigure}[b]{\linewidth}
        \includegraphics[width=\linewidth]{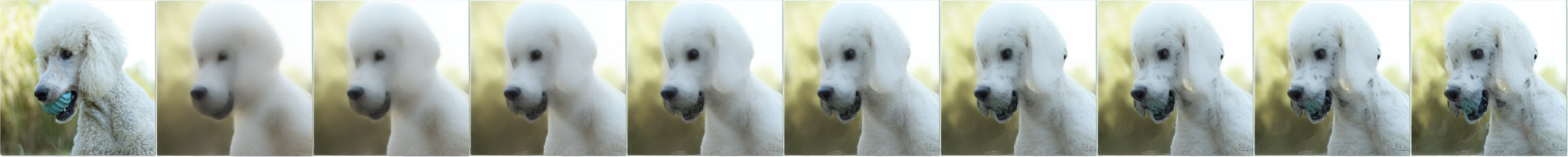}
        \caption{standard poodle $\rightarrow$ dalmatian}
    \end{subfigure}
    \begin{subfigure}[b]{\linewidth}
        \includegraphics[width=\linewidth]{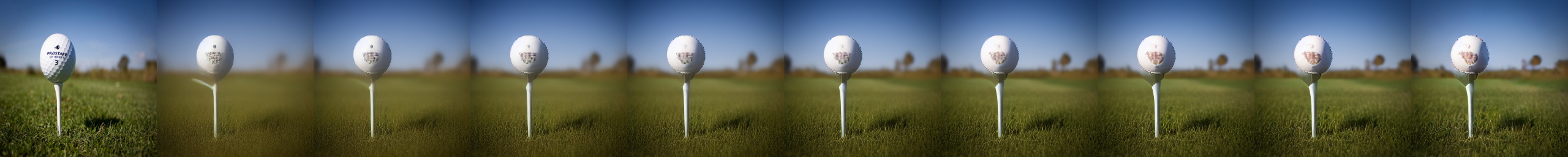}
        \caption{golf ball $\rightarrow$ baseball}
    \end{subfigure}
    \begin{subfigure}[b]{\linewidth}
        \includegraphics[width=\linewidth]{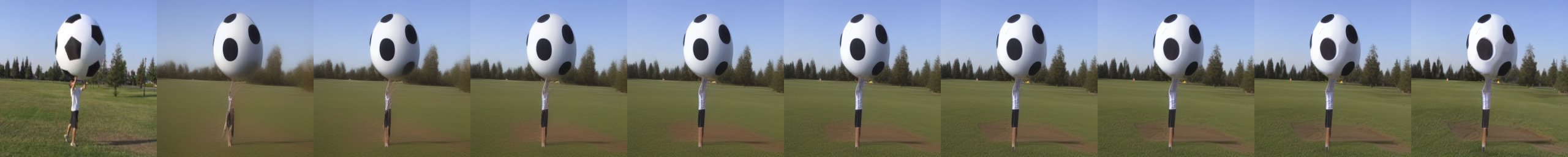}
        \caption{soccer ball $\rightarrow$ golf ball}
    \end{subfigure}
    \begin{subfigure}[b]{\linewidth}
        \includegraphics[width=\linewidth]{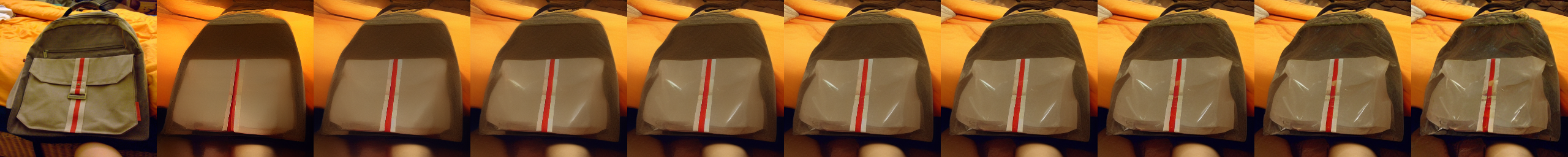}
        \caption{backpack $\rightarrow$ plastic bag}
    \end{subfigure}
    \begin{subfigure}[b]{\linewidth}
        \includegraphics[width=\linewidth]{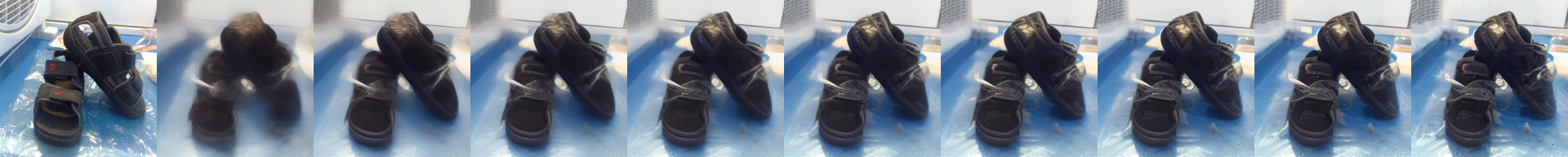}
        \caption{sandal $\rightarrow$ running shoe}
    \end{subfigure}
    \caption{Visualization of changes over the course of the counterfactual generation. From left to right: (f)actual image \begin{math}x^{\mathrm{F}}\end{math}, intermediate visualizations (linearly spaced) till final counterfactual \begin{math}x^{\mathrm{CF}}\end{math}.}
    \label{fig:evolution}
\end{figure}

We conducted a deeper analysis to understand how a (f)actual image \begin{math}x^{\mathrm{F}}\end{math} is transformed into a counterfactual explanation \begin{math}x^{\mathrm{CF}}\end{math}. To this end, we visualized intermediate steps (linearly spaced) of the diffusion process in \figref{fig:evolution}. We found that the image gradually evolves from \begin{math}x^{\mathrm{F}}\end{math} to \begin{math}x^{\mathrm{CF}}\end{math} by modifying coarse (low-frequency) features (\eg, blobs or shapes) in the earlier steps and more intricate (high-frequency) features (\eg, textures) in the latter steps of the diffusion process.

\section{Diversity of the generated counterfactual explanations}\label{sec:diversity}
\begin{figure}[t]
    \centering
    \begin{subfigure}[b]{\linewidth}
        \includegraphics[width=\linewidth]{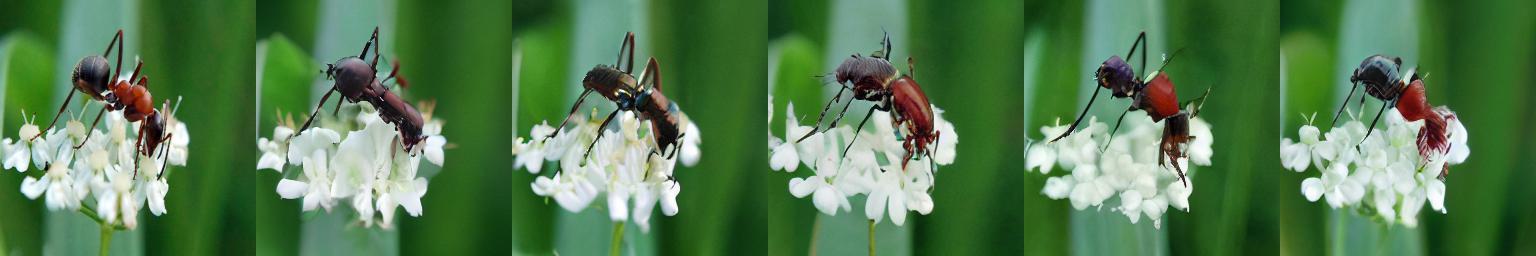}
        \caption{ant $\rightarrow$ ground beetle}
    \end{subfigure}
    \begin{subfigure}[b]{\linewidth}
        \includegraphics[width=\linewidth]{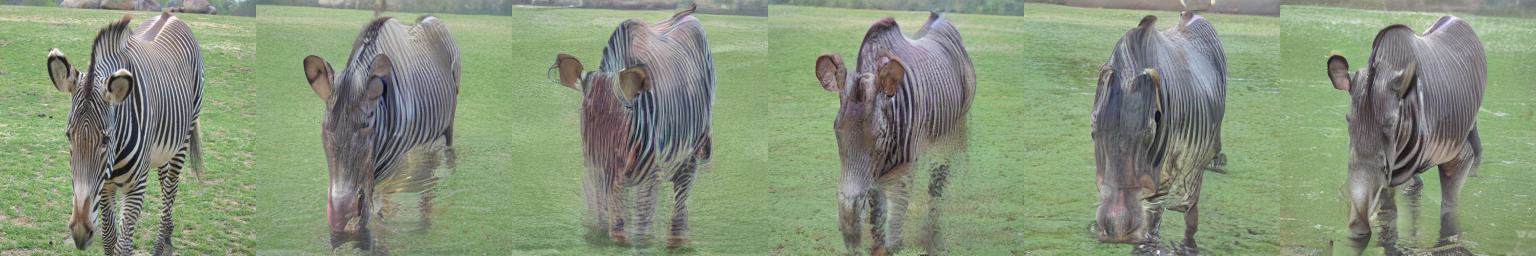}
        \caption{zebra $\rightarrow$ hippopotamus}
    \end{subfigure}

    \begin{subfigure}[b]{\linewidth}
        \includegraphics[width=\linewidth]{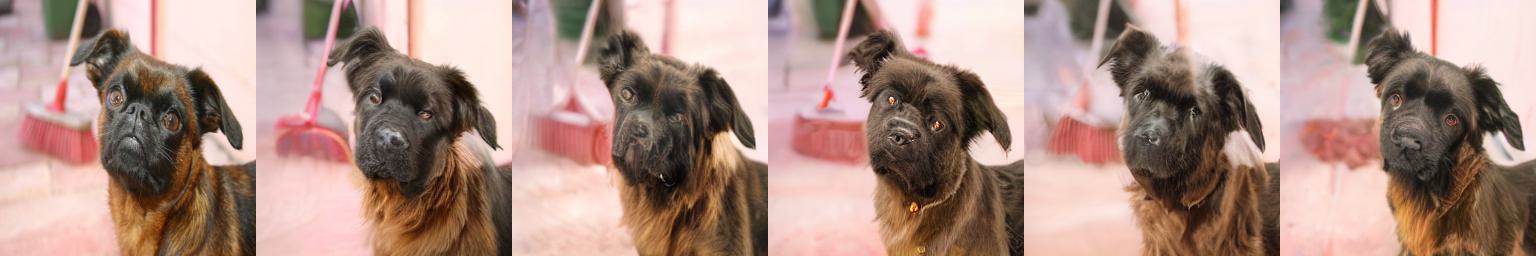}
        \caption{brabancon griffon $\rightarrow$ newfoundland}
    \end{subfigure}

    \begin{subfigure}[b]{\linewidth}
        \includegraphics[width=\linewidth]{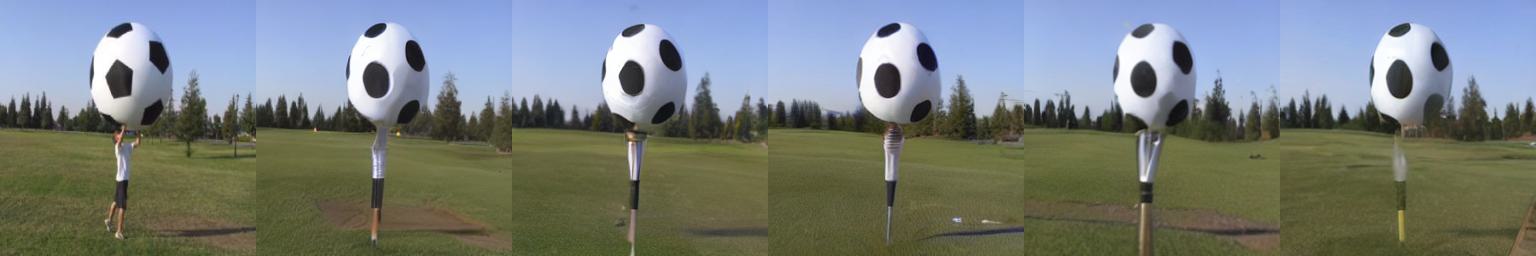}
        \caption{soccer ball $\rightarrow$ golfball}
    \end{subfigure}
        \begin{subfigure}[b]{\linewidth}
        \includegraphics[width=\linewidth]{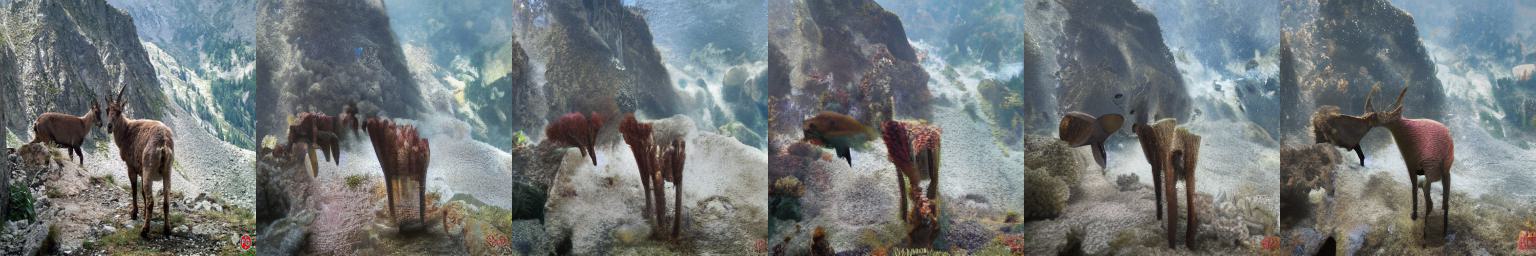}
        \caption{alp $\rightarrow$ coral reef}
    \end{subfigure}

    \caption{Qualitative diversity assessment across five different random seeds (0-4) using \ourst~on ImageNet \citep{deng2009imagenet} with ResNet-50 \citep{he2016deep}. From left to right: original image, counterfactual image generated by \ourst~for five different seeds.}
    \label{fig:diversity}
\end{figure}
Diffusion models by design are capable of generating image distributions. While the used DDIM sampler \citep{song2021denoising} is deterministic, we remark that the abduction step (application of forward diffusion onto the (f)actual input \begin{math}x^\textrm{F}\end{math}) still introduces stochasticity in our approach, resulting in the generation of diverse counterfactual images. More specifically, \figref{fig:diversity} shows that the injected noise influences the features that are added to or removed from the (f)actual image at different scales. Therefore, to gain a more comprehensive understanding of the underlying semantics driving the transitions in classifiers' decisions, we recommend to generate counterfactuals for multiple random seeds.

\section{Quantitative results on CelebA HQ}\label{sec:celeb_results}
\tabref{tab:celebahq} provides quantitative comparison to previous methods on CelebA HQ \citep{lee2020maskgan}. We compared to DiVE \citep{rodriguez2021beyond}, STEEX \citep{jacob2022steex}, DiME \citep{jeanneret2022diffusion}, and ACE \citep{jeanneret2023adversarial} on CelebA HQ \citep{lee2020maskgan} using a DenseNet-121 \citep{jacob2022steex}. Note that in contrast to previous works, \ourst~is not specifically trained on a face image distribution and still yields competitive quantitative results.
\tabref{tab:celebahq}, \figref{fig:other_datasets}(a)-(d) as well as \figref{fig:celeb_appendix} demonstrate the ability of \ourst~to capture and manipulate distinctive facial features, also showcasing its efficacy in the domain of human faces: \ourst~inserts or removes local features such as wrinkles, dimples, and eye bags when moving along the smile and age attributes. 
\begin{table*}[t]
    \centering
    \caption{Quantitative comparison to previous works using DenseNet-121 on CelebA HQ. All methods except for ours require access to the target classifier's training data distribution.}
    \resizebox{\linewidth}{!}{
    \begin{tabular}{cccccccc} \toprule
        \multicolumn{8}{c}{\textbf{Smile}}\\\midrule
        Method        & FID (\begin{math}\downarrow\end{math})  & sFID (\begin{math}\downarrow\end{math}) & FVA (\begin{math}\uparrow\end{math})  & FS (\begin{math}\uparrow\end{math}) & MNAC (\begin{math}\downarrow\end{math})  & CD (\begin{math}\downarrow\end{math}) & COUT (\begin{math}\uparrow\end{math})  \\ \midrule 
        DiVE \citep{rodriguez2021beyond} &107.0 & -    & 35.7 & -  & 7.41  & -     & -  \\ 
        STEEX \citep{jacob2022steex}    & 21.9 & -    & 97.6 & -         & 5.27  & -     & - \\
        DiME \citep{jeanneret2022diffusion} & 18.1 & 27.7 & 96.7 & 0.6729    & 2.63  & 1.82  & \textbf{0.6495} \\
        ACE \begin{math}\ell_1\end{math} \citep{jeanneret2023adversarial} & \textbf{3.21} & \textbf{20.2} & \textbf{100.0}& \textbf{0.8941} & \textbf{1.56}  & 2.61  & 0.5496 \\
        ACE \begin{math}\ell_2\end{math} \citep{jeanneret2023adversarial} & 6.93 & 22.0 & \textbf{100.0}& 0.8440    & 1.87  & 2.21  & 0.5946\\\midrule
        \ourst  & 13.6 & 25.8 & 99.1 & 0.756 & 2.44  & \textbf{1.68} &  0.3428 \\\midrule
        \multicolumn{8}{c}{\textbf{Age}}\\\midrule
        DiVE \citep{rodriguez2021beyond} & 107.5 & -    & 32.3 & -  & 6.76 & -    & - \\ 
        STEEX \citep{jacob2022steex} & 26.8 & -    & 96.0 & -         & 5.63 & -    & - \\
        DiME \citep{jeanneret2022diffusion} & 18.7 & 27.8 & 95.0 & 0.6597    & 2.10 & 4.29 & \textbf{0.5615} \\
        ACE \begin{math}\ell_1\end{math} \citep{jeanneret2023adversarial} &\textbf{5.31} & \textbf{21.7} & \textbf{99.6} & \textbf{0.8085} & \textbf{1.53} & 5.4  & 0.3984 \\
        ACE \begin{math}\ell_2\end{math} \citep{jeanneret2023adversarial} & 16.4 & 28.2 & \textbf{99.6} & 0.7743    & 1.92 & 4.21 & 0.5303 \\\midrule
        \ourst & 14.2 & 25.6 &  98.0 & 0.7319 & 2.12  & \textbf{4.02} & 0.3297  \\\bottomrule
    \end{tabular}
    }
    \label{tab:celebahq}
\end{table*}

\section{Additional qualitative examples}\label{sec:more_examples}
Figures \ref{fig:imagenet_appendix}, \ref{fig:celeb_appendix}, \ref{fig:pets_appendix}, and \ref{fig:flowers_appendix} provide additional qualitative examples for ImageNet with ResNet-50, CelebA HQ with DenseNet-121, Oxford Pets with OpenCLIP VIT-B/32, or Oxford Flowers 102 with (frozen) DINO-VIT-S/8 with (trained) linear classifier, respectively. Note that, in contrast to standard image generation, editing or prompt-to-prompt tuning, we are interested in \emph{minimal} semantically meaningful changes to \emph{flip} a target classifier's prediction (and not just generating the best looking image).
\begin{figure}[ht]
    \centering
    \begin{subfigure}[b]{\linewidth}
        \includegraphics[width=\linewidth]{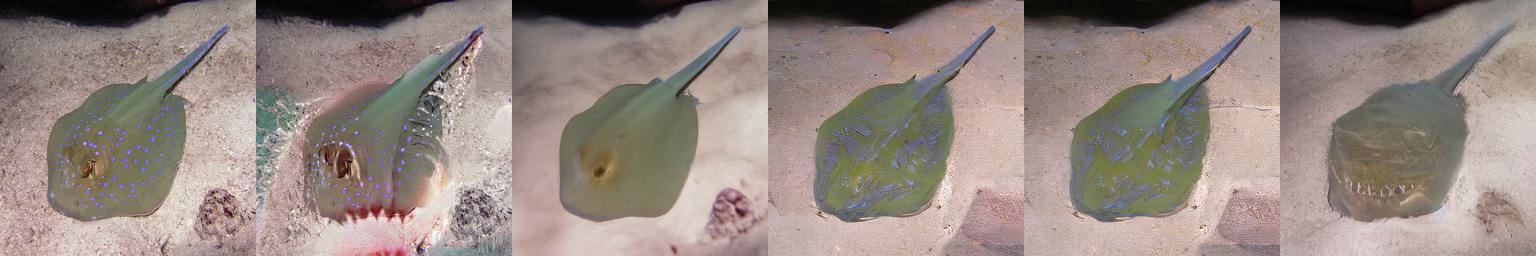}
        \caption{stingray $\rightarrow$ great white shark}
    \end{subfigure}
    \begin{subfigure}[b]{\linewidth}
        \includegraphics[width=\linewidth]{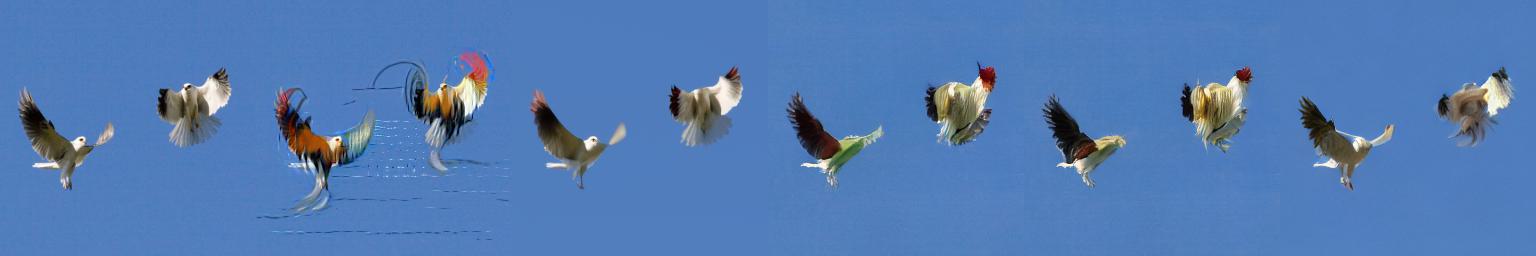}
        \caption{kite $\rightarrow$ rooster}
    \end{subfigure}
    \begin{subfigure}[b]{\linewidth}
        \includegraphics[width=\linewidth]{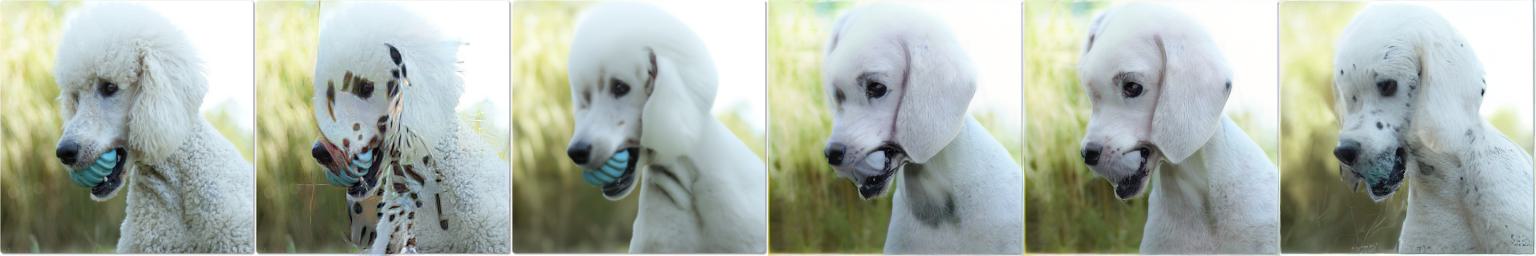}
        \caption{poodle $\rightarrow$ dalmatian}
    \end{subfigure}
    \begin{subfigure}[b]{\linewidth}
        \includegraphics[width=\linewidth]{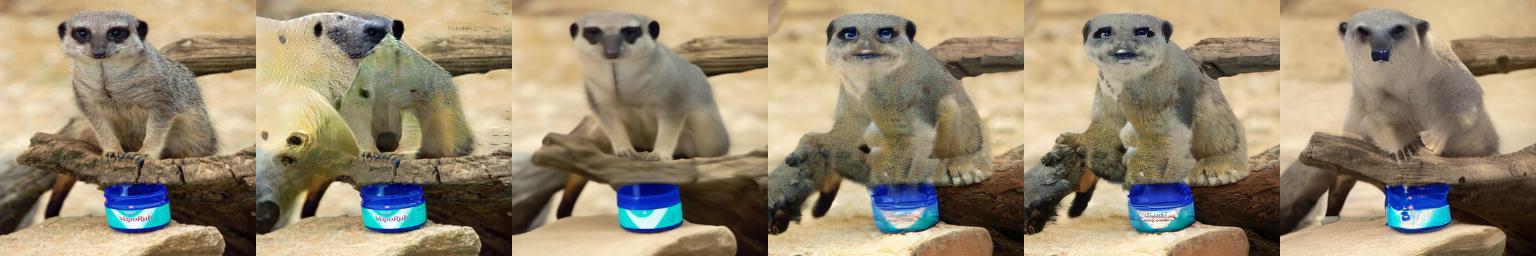}
        \caption{meerkat $\rightarrow$ ice bear}
    \end{subfigure}
    \begin{subfigure}[b]{\linewidth}
        \includegraphics[width=\linewidth]{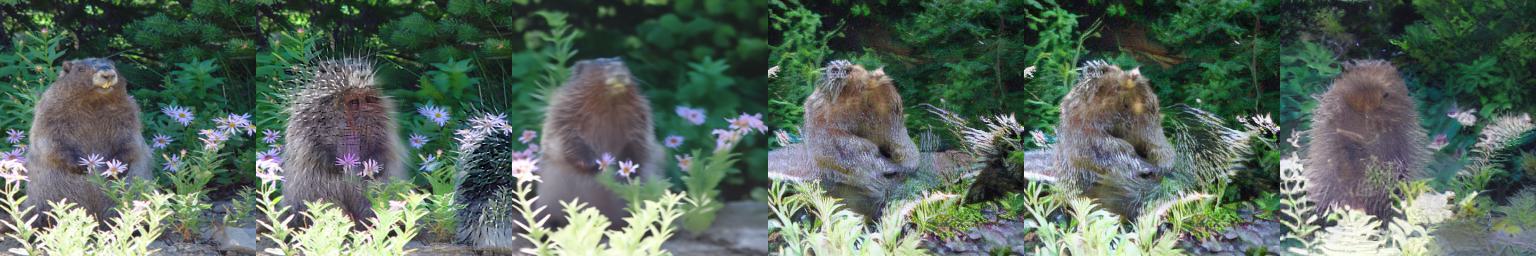}
        \caption{marmot $\rightarrow$ porcupine}
    \end{subfigure}
    \begin{subfigure}[b]{\linewidth}
        \includegraphics[width=\linewidth]{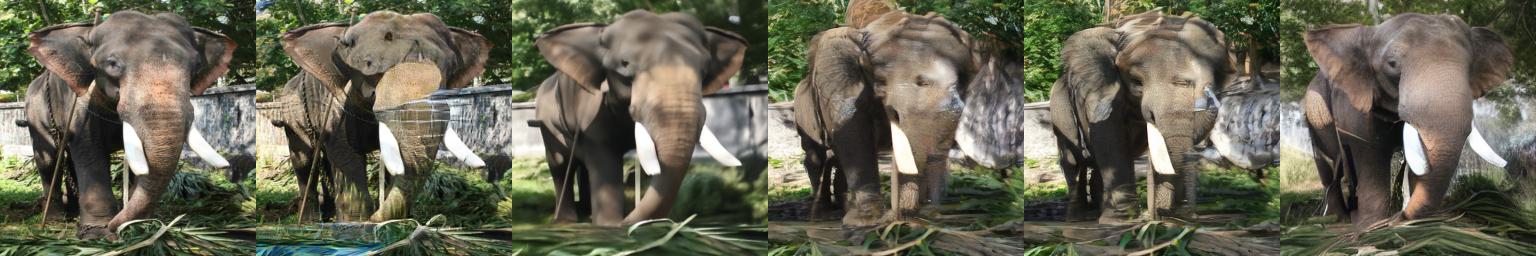}
        \caption{indian elephant $\rightarrow$ african elephant}
    \end{subfigure}
    \begin{subfigure}[b]{\linewidth}
        \includegraphics[width=\linewidth]{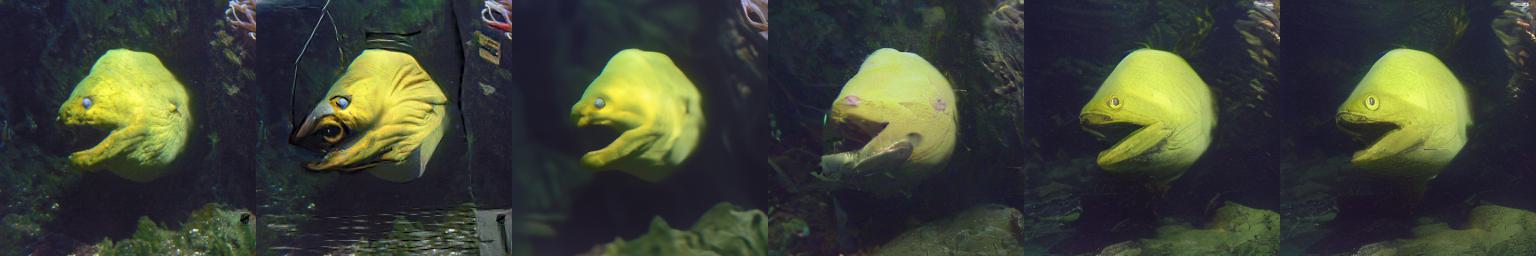}
        \caption{eel $\rightarrow$ coho}
    \end{subfigure}
\end{figure}
\begin{figure}
    \ContinuedFloat
    \begin{subfigure}[b]{\linewidth}
        \includegraphics[width=\linewidth]{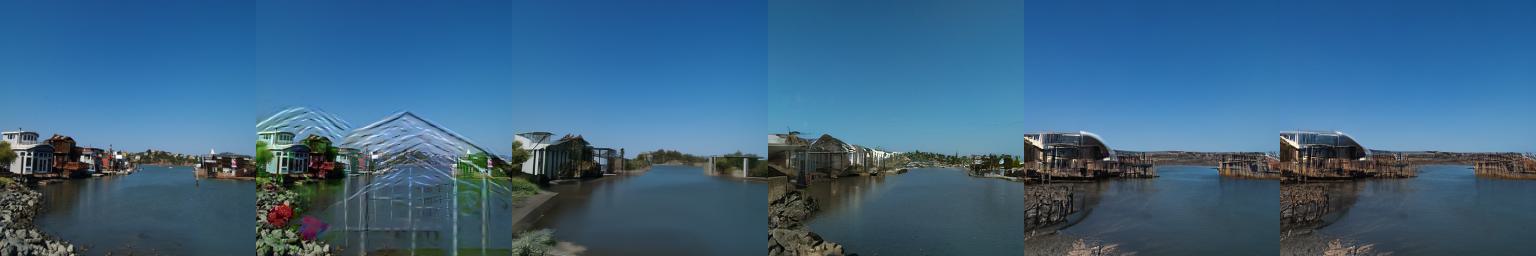}
        \caption{boathouse $\rightarrow$ greenhouse}
    \end{subfigure}
    \begin{subfigure}[b]{\linewidth}
        \includegraphics[width=\linewidth]{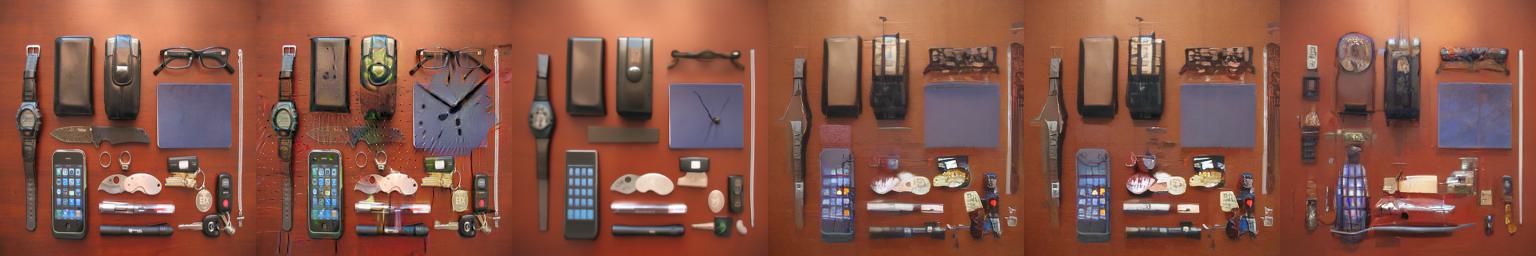}
        \caption{digital watch $\rightarrow$ wall clock}
    \end{subfigure}
    \begin{subfigure}[b]{\linewidth}
        \includegraphics[width=\linewidth]{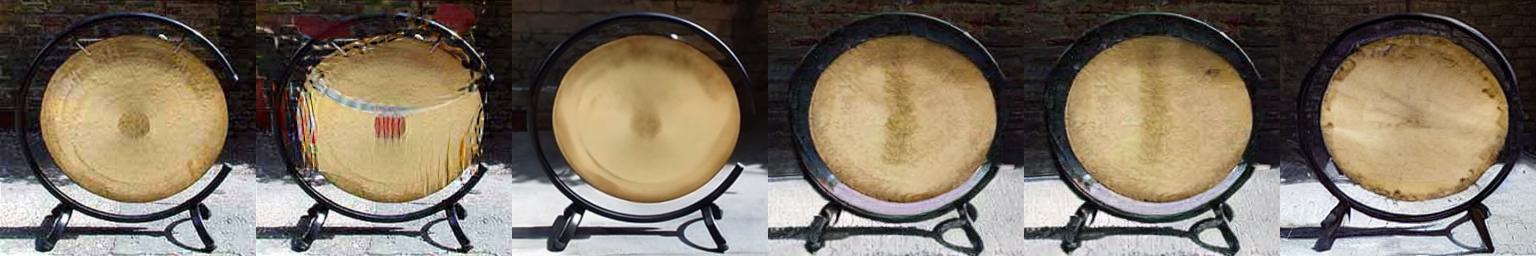}
        \caption{gong $\rightarrow$ drum}
    \end{subfigure}
    \begin{subfigure}[b]{\linewidth}
        \includegraphics[width=\linewidth]{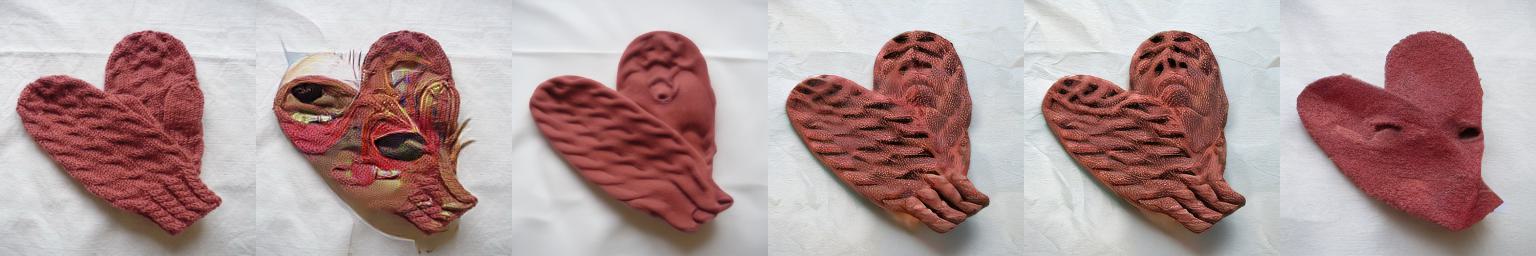}
        \caption{mitten $\rightarrow$ mask}
    \end{subfigure}
    \begin{subfigure}[b]{\linewidth}
        \includegraphics[width=\linewidth]{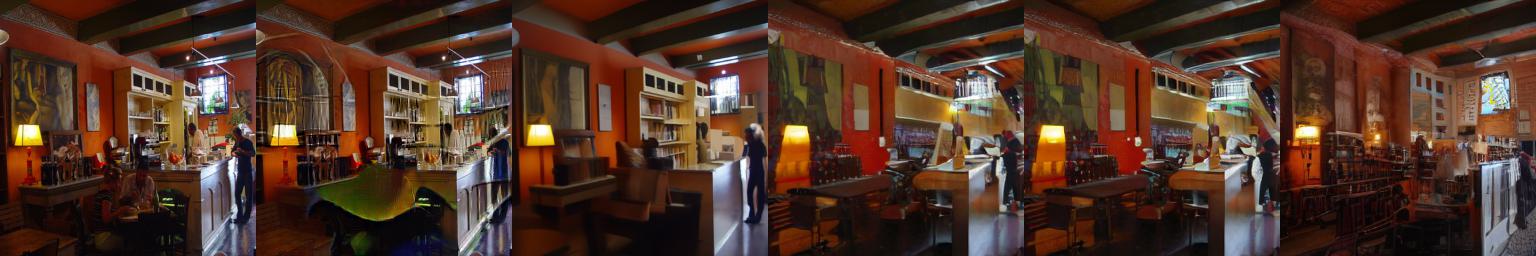}
        \caption{restaurant $\rightarrow$ library}
    \end{subfigure}
    \begin{subfigure}[b]{\linewidth}
        \includegraphics[width=\linewidth]{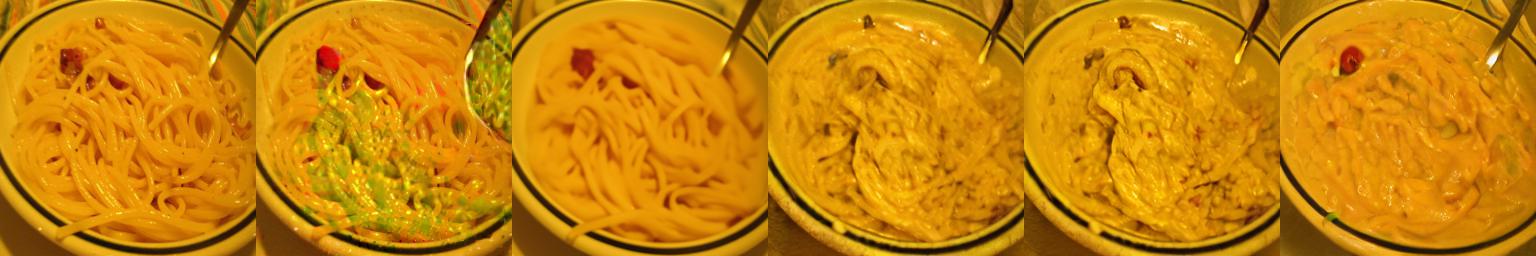}
        \caption{carbonara $\rightarrow$ guacamole}
    \end{subfigure}
    \begin{subfigure}[b]{\linewidth}
        \includegraphics[width=\linewidth]{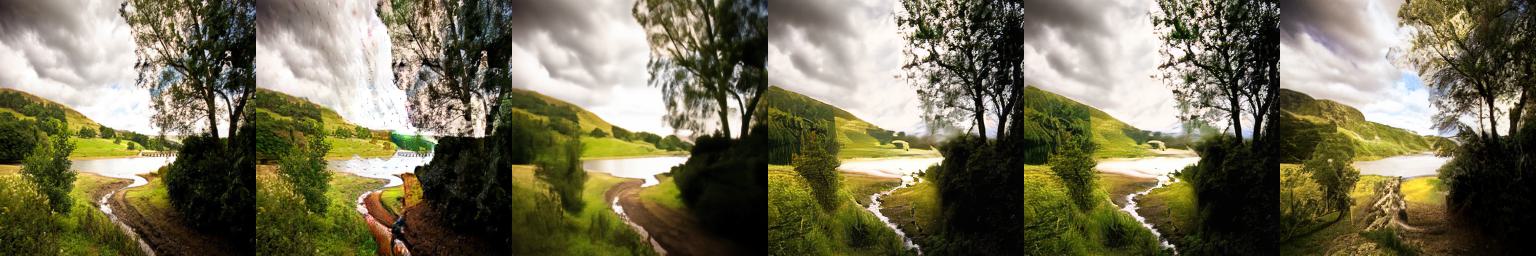}
        \caption{valley $\rightarrow$ cliff}
    \end{subfigure}
    \caption{Additional qualitative results for on ImageNet with ResNet-50. From left to right: original image, counterfactual images generated by SVCE \citep{boreiko2022sparse}, DVCE \citep{augustin2022diffusion}, LDCE-no consensus, \oursc, and \ourst.}
    \label{fig:imagenet_appendix}
\end{figure}
\begin{figure}[ht]
    \centering
    \begin{subfigure}[b]{\linewidth}
        \includegraphics[width=\linewidth]{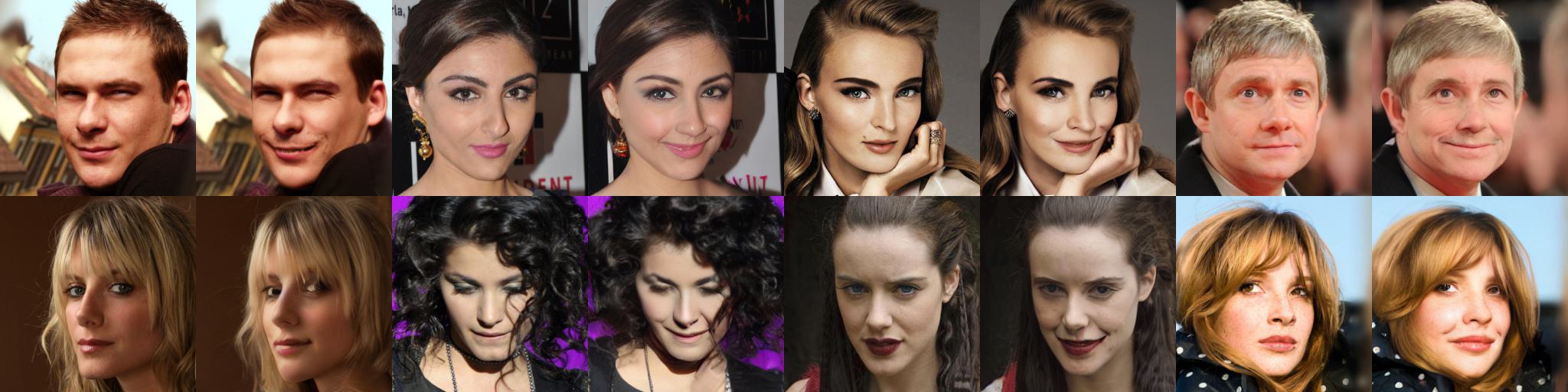}
        \caption{no-smile $\rightarrow$ smile}
    \end{subfigure}
    \begin{subfigure}[b]{\linewidth}
        \includegraphics[width=\linewidth]{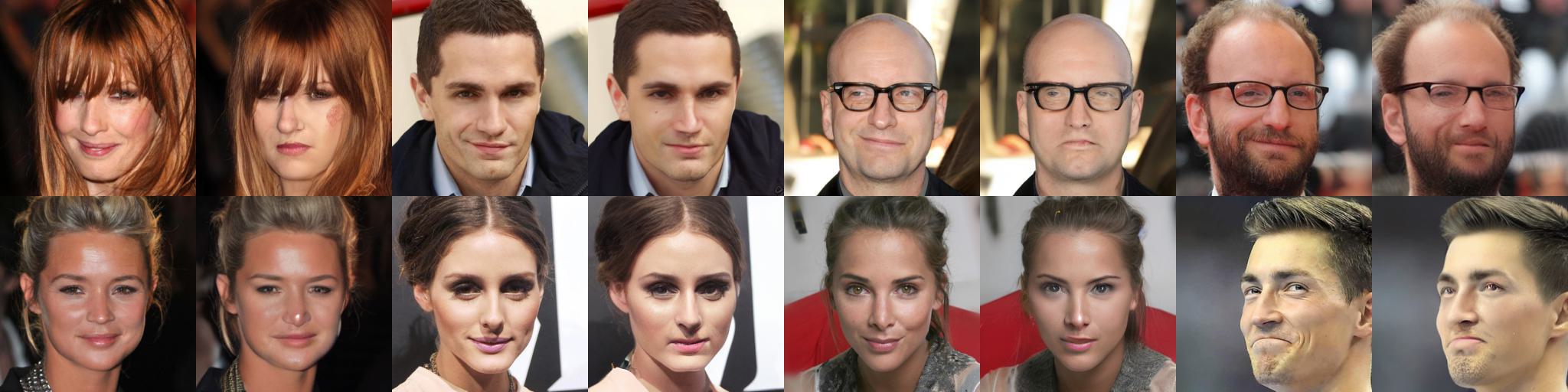}
        \caption{smile $\rightarrow$ no-smile}
    \end{subfigure}
    \begin{subfigure}[b]{\linewidth}
        \includegraphics[width=\linewidth]{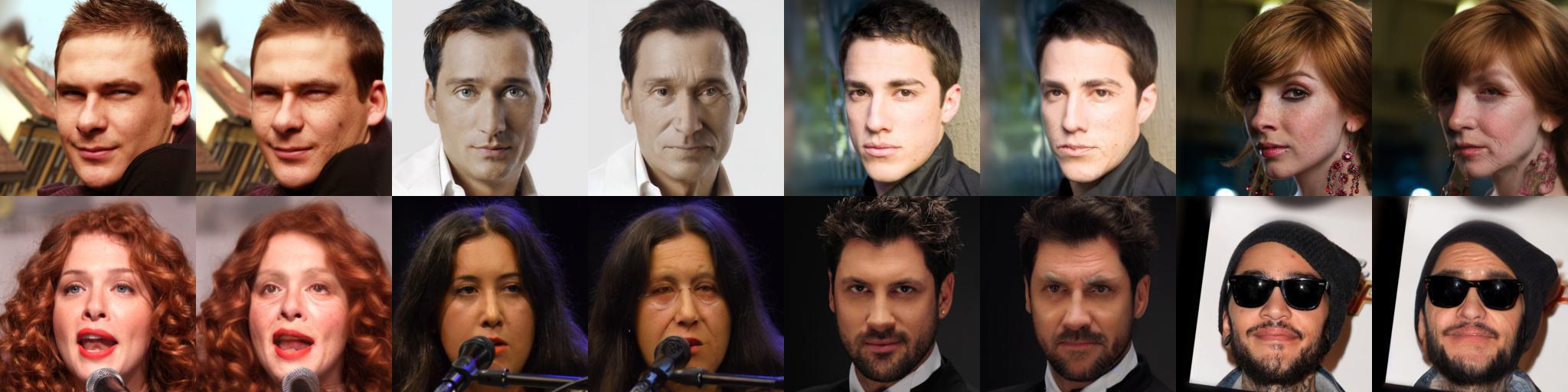}
        \caption{young $\rightarrow$ old}
    \end{subfigure}
    \begin{subfigure}[b]{\linewidth}
        \includegraphics[width=\linewidth]{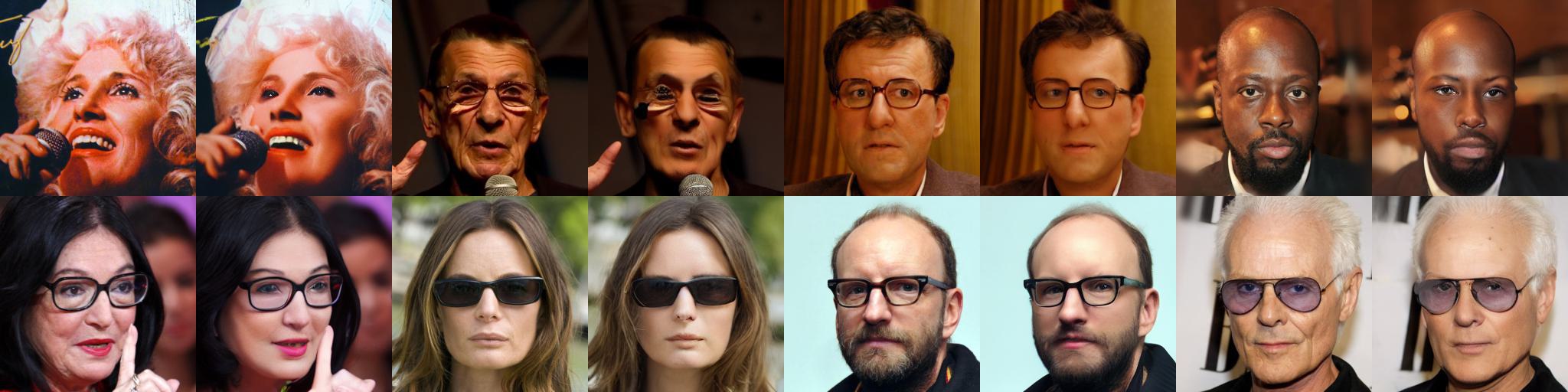}
        \caption{old $\rightarrow$ young}
    \end{subfigure}
    \caption{Additional qualitative results for \ourst~on CelebA HQ with DenseNet-121. Left: original image. Right: counterfactual image.}
    \label{fig:celeb_appendix}
\end{figure}
\begin{figure}[ht]
    \centering
    \begin{subfigure}[b]{0.32\linewidth}
        \includegraphics[width=\linewidth]{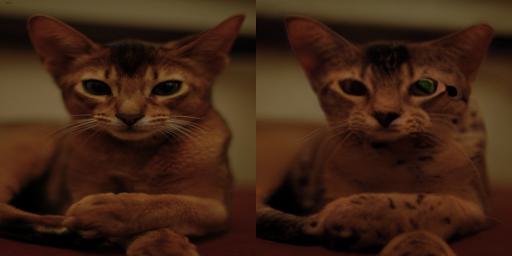}
        \caption{abyssinian $\rightarrow$ egyptian}
    \end{subfigure}
    \begin{subfigure}[b]{0.32\linewidth}
        \includegraphics[width=\linewidth]{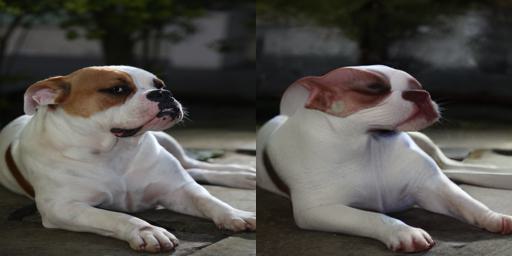}
        \caption{american bulldog $\rightarrow$ sphynx}
    \end{subfigure}
    \begin{subfigure}[b]{0.32\linewidth}
        \includegraphics[width=\linewidth]{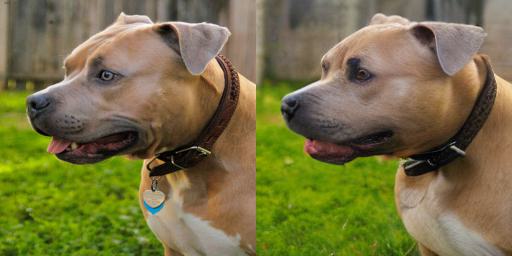}
        \caption{american pitbull $\rightarrow$ stafford}
    \end{subfigure}
    \begin{subfigure}[b]{0.32\linewidth}
        \includegraphics[width=\linewidth]{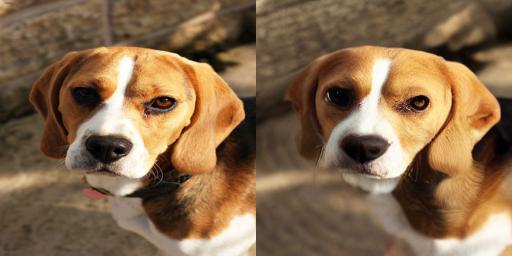}
        \caption{beagle $\rightarrow$ chihuahua}
    \end{subfigure}
    \begin{subfigure}[b]{0.32\linewidth}
        \includegraphics[width=\linewidth]{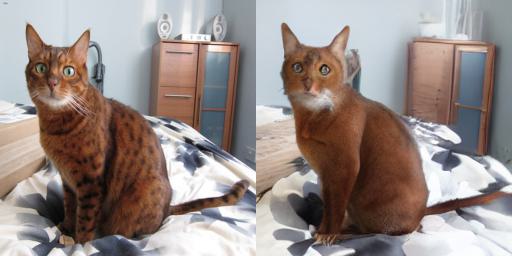}
        \caption{bengal $\rightarrow$ abyssinian}
    \end{subfigure}
    \begin{subfigure}[b]{0.32\linewidth}
        \includegraphics[width=\linewidth]{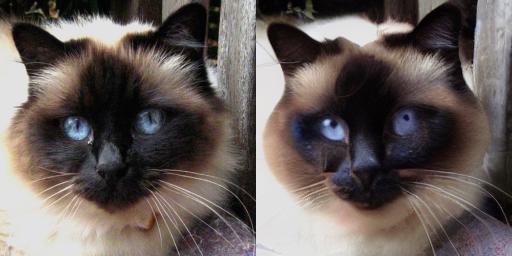}
        \caption{birman $\rightarrow$ siamese}
    \end{subfigure}
    \begin{subfigure}[b]{0.32\linewidth}
        \includegraphics[width=\linewidth]{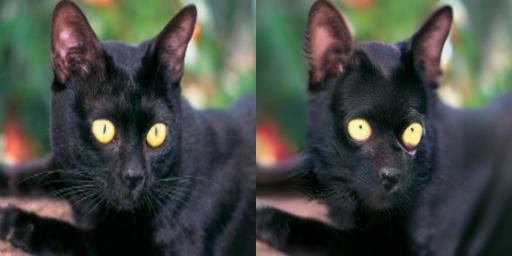}
        \caption{bombay $\rightarrow$ chihuahua}
    \end{subfigure}
    \begin{subfigure}[b]{0.32\linewidth}
        \includegraphics[width=\linewidth]{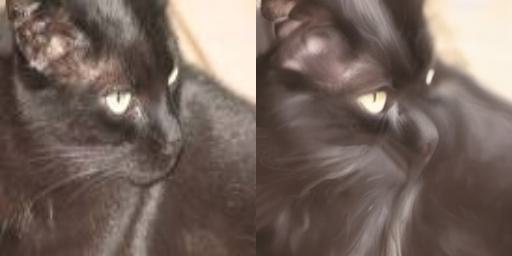}
        \caption{bombay $\rightarrow$ persian}
    \end{subfigure}
    \begin{subfigure}[b]{0.32\linewidth}
        \includegraphics[width=\linewidth]{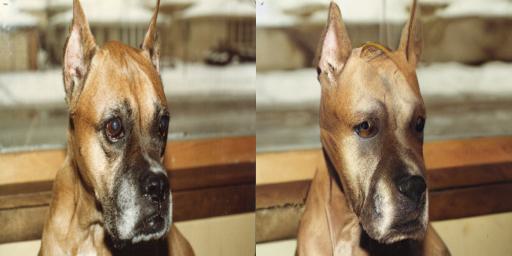}
        \caption{boxer $\rightarrow$ american pitbull}
    \end{subfigure}
    \begin{subfigure}[b]{0.32\linewidth}
        \includegraphics[width=\linewidth]{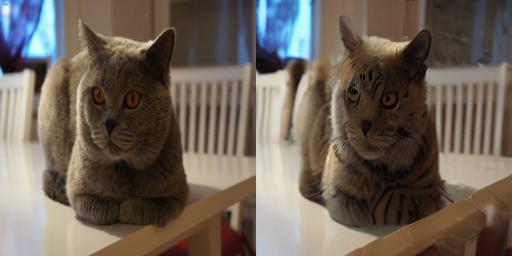}
        \caption{british shorthair $\rightarrow$ bengal}
    \end{subfigure}
    \begin{subfigure}[b]{0.32\linewidth}
        \includegraphics[width=\linewidth]{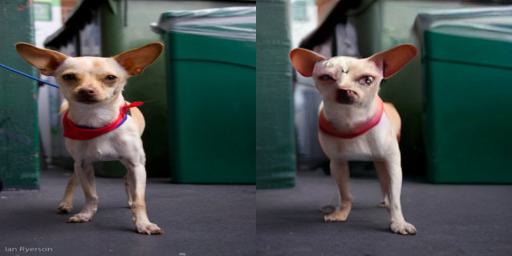}
        \caption{chihuahua $\rightarrow$ sphynx}
    \end{subfigure}
    \begin{subfigure}[b]{0.32\linewidth}
        \includegraphics[width=\linewidth]{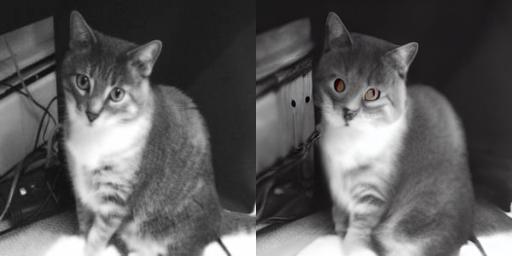}
        \caption{egyptian $\rightarrow$ british shorthair}
    \end{subfigure}
    \begin{subfigure}[b]{0.32\linewidth}
        \includegraphics[width=\linewidth]{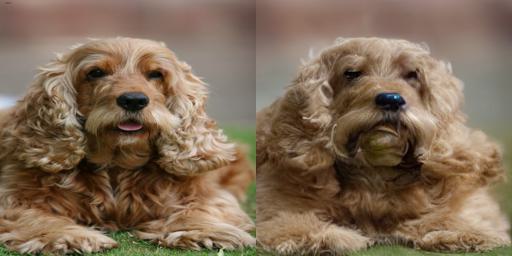}
        \caption{spaniel $\rightarrow$ wheaten}
    \end{subfigure}
    \begin{subfigure}[b]{0.32\linewidth}
        \includegraphics[width=\linewidth]{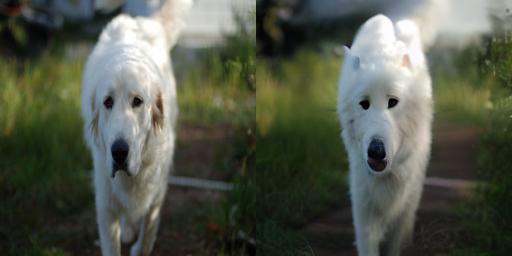}
        \caption{great pyrenees $\rightarrow$ samoyed}
    \end{subfigure}
    \begin{subfigure}[b]{0.32\linewidth}
        \includegraphics[width=\linewidth]{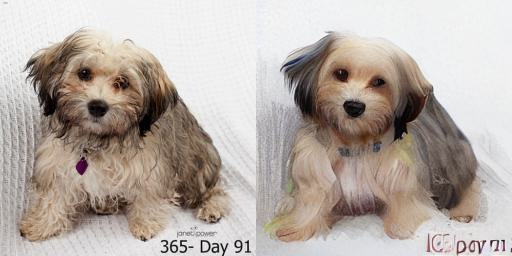}
        \caption{havanese $\rightarrow$ yorkshire}
    \end{subfigure}
    \begin{subfigure}[b]{0.32\linewidth}
        \includegraphics[width=\linewidth]{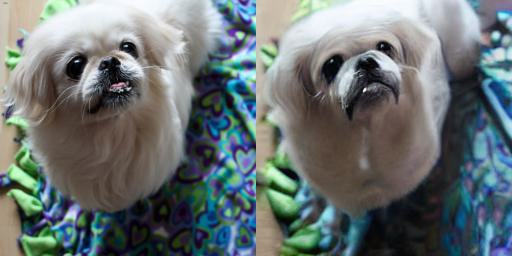}
        \caption{japanese chin $\rightarrow$ boxer}
    \end{subfigure}
    \begin{subfigure}[b]{0.32\linewidth}
        \includegraphics[width=\linewidth]{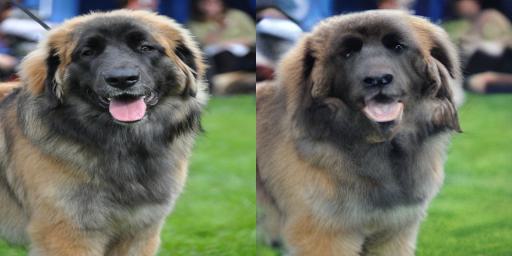}
        \caption{leonberger $\rightarrow$ newfoundland}
    \end{subfigure}
    \begin{subfigure}[b]{0.32\linewidth}
        \includegraphics[width=\linewidth]{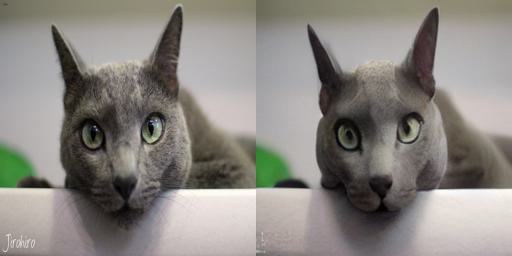}
        \caption{russian blue $\rightarrow$ sphynx}
    \end{subfigure}
    \begin{subfigure}[b]{0.32\linewidth}
        \includegraphics[width=\linewidth]{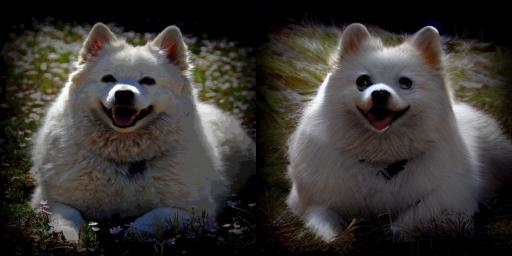}
        \caption{samoyed $\rightarrow$ pomeranian}
    \end{subfigure}
    \begin{subfigure}[b]{0.32\linewidth}
        \includegraphics[width=\linewidth]{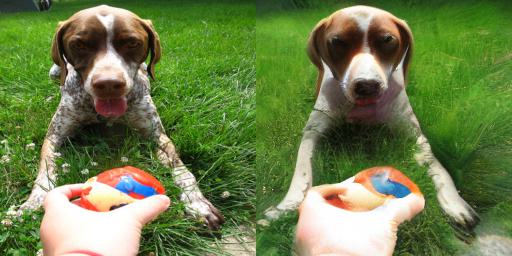}
        \caption{german shorthaired $\rightarrow$ beagle}
    \end{subfigure}
    \begin{subfigure}[b]{0.32\linewidth}
        \includegraphics[width=\linewidth]{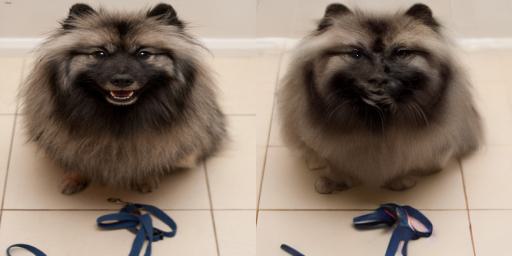}
        \caption{keeshond $\rightarrow$ birman}
    \end{subfigure}
    \caption{Additional qualitative results for \ourst~on Oxford Pets with OpenCLIP VIT-B/32. Left: original image. Right: counterfactual image.}
    \label{fig:pets_appendix}
\end{figure}
\begin{figure}[ht]
    \centering
    \begin{subfigure}[b]{0.32\linewidth}
        \includegraphics[width=\linewidth]{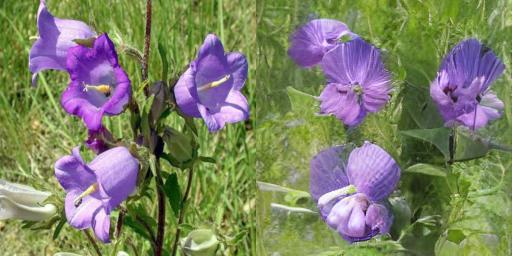}
        \caption{canterbury $\rightarrow$ balloon flower}
    \end{subfigure}
    \begin{subfigure}[b]{0.32\linewidth}
        \includegraphics[width=\linewidth]{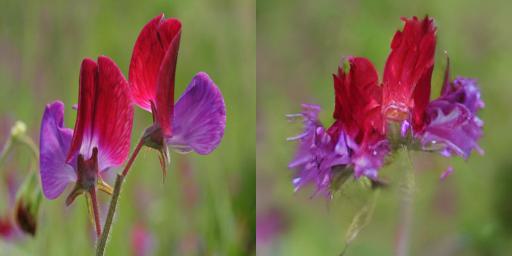}
        \caption{sweet pea $\rightarrow$ bee balm}
    \end{subfigure}
    \begin{subfigure}[b]{0.32\linewidth}
        \includegraphics[width=\linewidth]{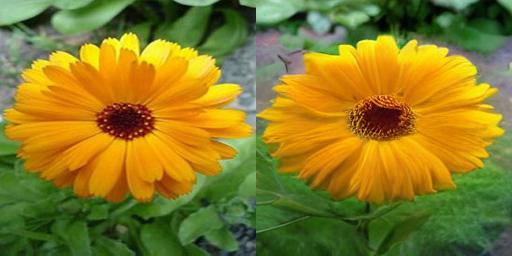}
        \caption{english marigold $\rightarrow$ sunflower}
    \end{subfigure}
    \begin{subfigure}[b]{0.32\linewidth}
        \includegraphics[width=\linewidth]{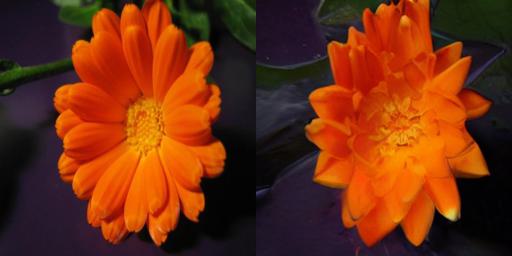}
        \caption{english marigold $\rightarrow$ water lily}
    \end{subfigure}
    \begin{subfigure}[b]{0.32\linewidth}
        \includegraphics[width=\linewidth]{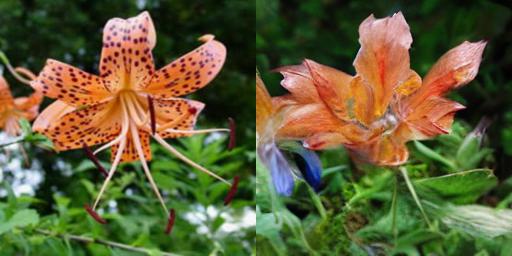}
        \caption{tiger lily $\rightarrow$ stemless gentian}
    \end{subfigure}
    \begin{subfigure}[b]{0.32\linewidth}
        \includegraphics[width=\linewidth]{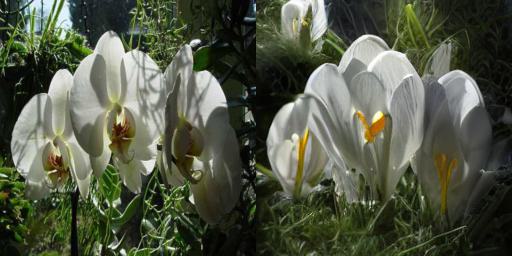}
        \caption{moon orchid $\rightarrow$ spring crocus}
    \end{subfigure}
    \begin{subfigure}[b]{0.32\linewidth}
        \includegraphics[width=\linewidth]{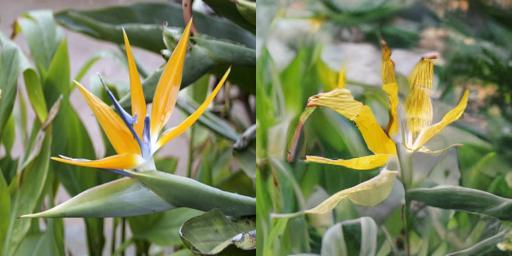}
        \caption{bird of paradise $\rightarrow$ yellow iris}
    \end{subfigure}
    \begin{subfigure}[b]{0.32\linewidth}
        \includegraphics[width=\linewidth]{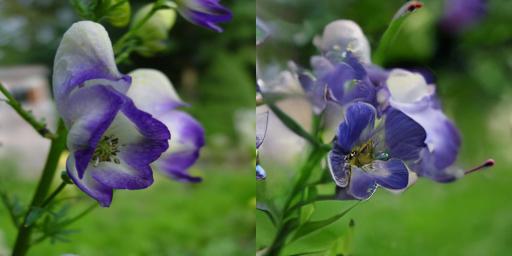}
        \caption{monkshood $\rightarrow$ wallflower}
    \end{subfigure}
    \begin{subfigure}[b]{0.32\linewidth}
        \includegraphics[width=\linewidth]{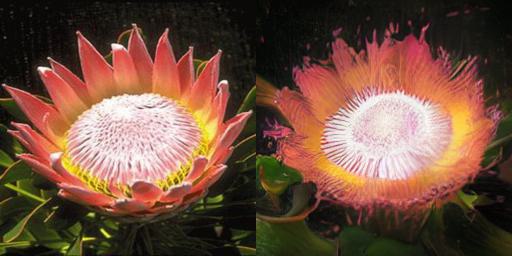}
        \caption{king protea $\rightarrow$ passion flower}
    \end{subfigure}
    \begin{subfigure}[b]{0.32\linewidth}
        \includegraphics[width=\linewidth]{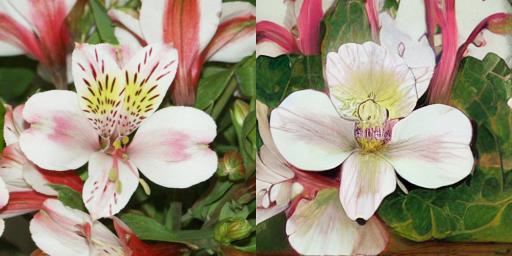}
        \caption{preuvian lily $\rightarrow$ watercress}
    \end{subfigure}
    \begin{subfigure}[b]{0.32\linewidth}
        \includegraphics[width=\linewidth]{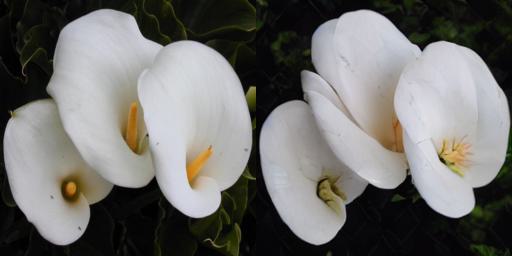}
        \caption{arum lily $\rightarrow$ magnolia}
    \end{subfigure}
    \begin{subfigure}[b]{0.32\linewidth}
        \includegraphics[width=\linewidth]{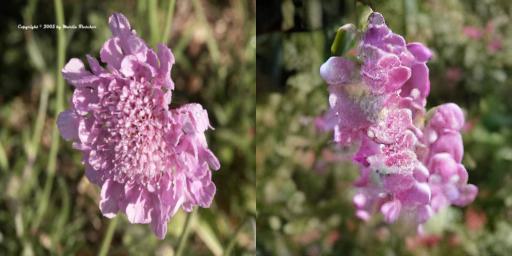}
        \caption{pincushion $\rightarrow$ snapdragon}
    \end{subfigure}
    \begin{subfigure}[b]{0.32\linewidth}
        \includegraphics[width=\linewidth]{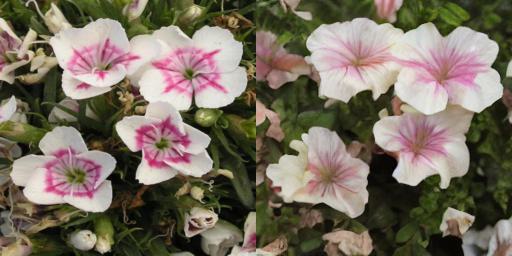}
        \caption{sweet william $\rightarrow$ petunia}
    \end{subfigure}
    \begin{subfigure}[b]{0.32\linewidth}
        \includegraphics[width=\linewidth]{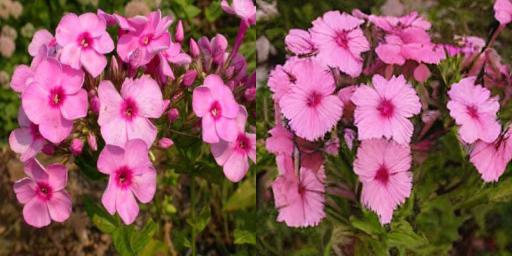}
        \caption{garden phlox $\rightarrow$ sweet william}
    \end{subfigure}
    \begin{subfigure}[b]{0.32\linewidth}
        \includegraphics[width=\linewidth]{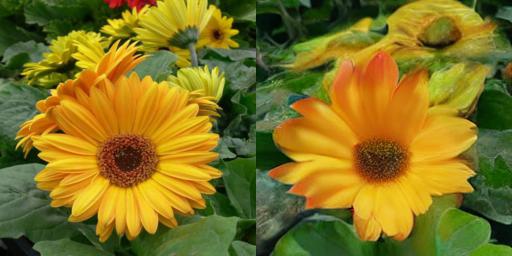}
        \caption{barbeton daisy $\rightarrow$ gazania}
    \end{subfigure}
    \begin{subfigure}[b]{0.32\linewidth}
        \includegraphics[width=\linewidth]{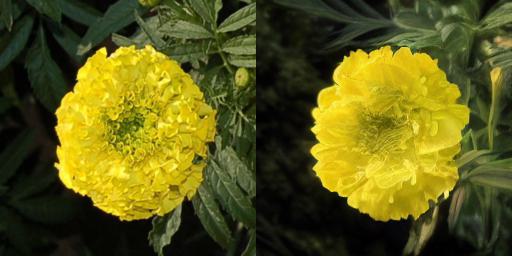}
        \caption{marigold $\rightarrow$ carnation}
    \end{subfigure}
    \begin{subfigure}[b]{0.32\linewidth}
        \includegraphics[width=\linewidth]{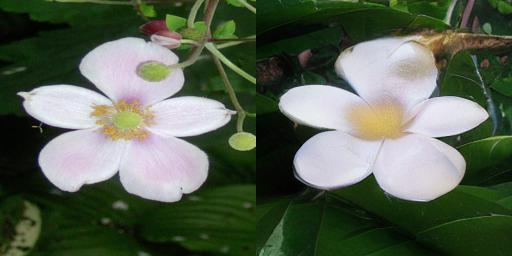}
        \caption{japanese anemone $\rightarrow$ frangipani}
    \end{subfigure}
    \begin{subfigure}[b]{0.32\linewidth}
        \includegraphics[width=\linewidth]{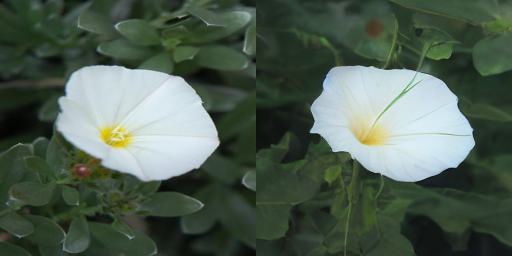}
        \caption{silverbush $\rightarrow$ morning glory}
    \end{subfigure}
    \begin{subfigure}[b]{0.32\linewidth}
        \includegraphics[width=\linewidth]{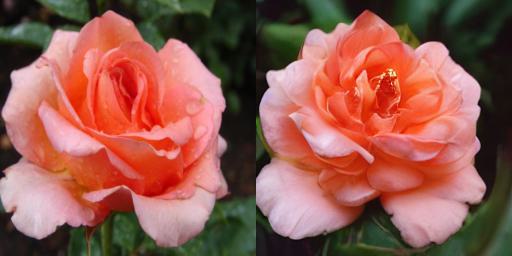}
        \caption{rose $\rightarrow$ camellia}
    \end{subfigure}
    \begin{subfigure}[b]{0.32\linewidth}
        \includegraphics[width=\linewidth]{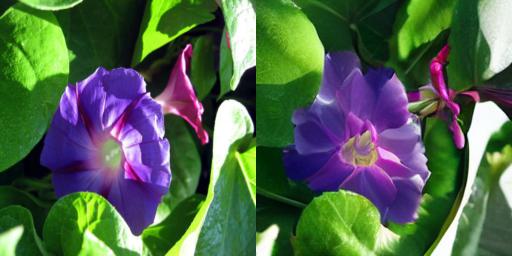}
        \caption{morning glory $\rightarrow$ desert-rose}
    \end{subfigure}
    \begin{subfigure}[b]{0.32\linewidth}
        \includegraphics[width=\linewidth]{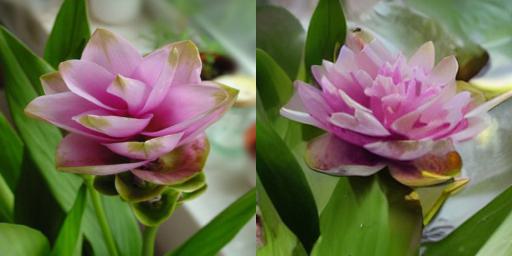}
        \caption{siam tulip $\rightarrow$ water lily}
    \end{subfigure}
    \caption{Additional qualitative results for \ourst~on Oxford Flowers 102 with (frozen) DINO-VIT-S/8 with (trained) linear classifier. Left: original image. Right: counterfactual image.}
    \label{fig:flowers_appendix}
\end{figure}

\section{Finetuning details}\label{sec:finetuning_errors}
We finetuned the final linear layer of ResNet-50 on the ImageNet training set combined with 25 examples that correspond to the respective model error type for 16 epochs and a batch size of 512. We use stochastic gradient descent with learning rate of 0.1, momentum of 0.9, and weight decay of 0.0005. We used cosine annealing as learning rate scheduler and standard image augmentations (random crop, horizontal flip, and normalization). We evaluated the final model on the holdout test set.

\clearpage
\section{Failure modes}\label{sec:failures}
\begin{figure}[t]
    \centering
    \begin{subfigure}[b]{0.4\linewidth}
        \includegraphics[width=\linewidth]{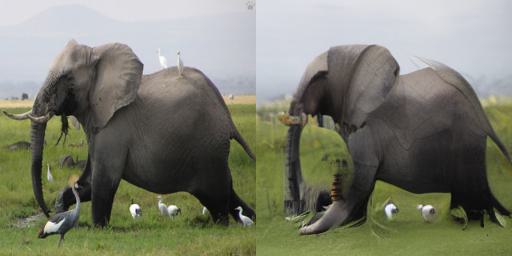}
        \caption{\centering bluriness\newline african elephant $\rightarrow$ armadillo\label{fig:failure_blurry}}
    \end{subfigure}
    \begin{subfigure}[b]{0.4\linewidth}
        \includegraphics[width=\linewidth]{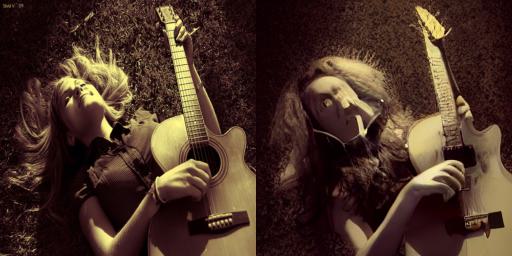}
        \caption{\centering distorted bodies\newline acoustic guitar $\rightarrow$ electric guitar\label{fig:failure_distored_bodies}}
    \end{subfigure}
    \begin{subfigure}[b]{0.4\linewidth}
        \includegraphics[width=\linewidth]{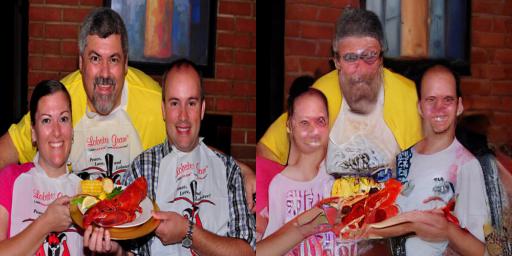}
        \caption{\centering distorted faces\newline american lobster $\rightarrow$ dungeness crab balm\label{fig:failure_distored_faces}}
    \end{subfigure}
    \begin{subfigure}[b]{0.4\linewidth}
        \includegraphics[width=\linewidth]{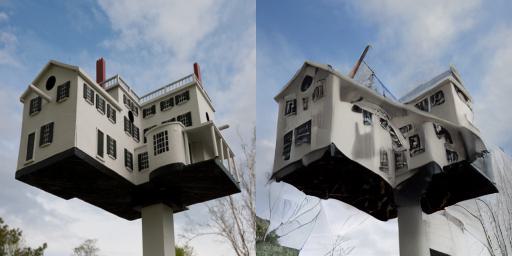}
        \caption{\centering large distance to counterfactual class \begin{math}y^{\textrm{CF}}\end{math}\newline birdhouse $\rightarrow$ umbrella\label{fig:failure_large_gap}}
    \end{subfigure}
    \caption{Failure modes of LDCE (\ie, \ourst) on ImageNet \citep{deng2009imagenet} with ResNet-50 \citep{he2016deep}. Left: original image. Right: counterfactual image.}
    \label{fig:failures}
\end{figure}

In this section, we aim to disclose some observed failure modes of LDCE (specifically \ourst): (i) occasional blurry images (\figref{fig:failure_blurry}), (ii) distorted human bodies and faces (\figref{fig:failure_distored_bodies} and \ref{fig:failure_distored_faces}), and (iii) a large distance to the counterfactual target class causing difficulties in counterfactual generation (\figref{fig:failure_large_gap}). Moreover, we note that these failure modes are further aggravated when multiple instances of the same class (\figref{fig:failure_distored_faces}) or multiple classes or objects are present in the image (\figref{fig:failure_blurry}). 

As discussed in our limitations section (\secref{sec:limitations}), we believe that the former cases (i \& ii) can mostly be attributed to limitations in the foundation diffusion model, which can potentially be addressed through orthogonal advancements in generative modeling. On the other hand, the latter case (iii) could potentially be overcome by further hyperparameters tuning, \eg, increasing  classifier strength \begin{math}\lambda_c\end{math} and decreasing the distance strength \begin{math}\lambda_d\end{math}. However, it is important to note that such adjustments may lead to counterfactuals that are farther away from the original instance, thereby possibly violating the desired desiderata of closeness. Another approach would be to increase the number of diffusion steps \begin{math}T\end{math}, but this would result in longer counterfactual generation times. Achieving a balance for these hyperparameters is highly dependent on the specific user requirements and the characteristics of the dataset.

\end{document}